\documentclass[pdflatex,sn-mathphys-num]{sn-jnl}% Math and Physical Sciences Numbered Reference Style
\usepackage{graphicx}%
\usepackage{multirow}%
\usepackage{amsmath,amssymb,amsfonts}%
\usepackage{amsthm}%
\usepackage{mathrsfs}%
\usepackage[title]{appendix}%
\usepackage{xcolor}%
\usepackage{textcomp}%
\usepackage{manyfoot}%
\usepackage{booktabs}%
\usepackage{algorithm}%
\usepackage{algorithmicx}%
\usepackage{algpseudocode}%
\usepackage{listings}%
\usepackage{makecell}
\usepackage{dsfont}
\theoremstyle{thmstyleone}%
\theoremstyle{thmstyletwo}%

\theoremstyle{thmstylethree}%

\begin{document}

\title[Article Title]{CTC: The Composite Task Challenge for Cooperative Multi-Agent Reinforcement Learning}


\author*[1]{\fnm{Yurui} \sur{Li}}\email{liyr@zju.edu.cn}

\author[1]{\fnm{Yuxuan} \sur{Chen}}\email{yuxuan\_chen@zju.edu.cn}
\author[1]{\fnm{Xiaoli} \sur{Yang}}\email{xiaoliyang@zju.edu.cn}
\author*[1]{\fnm{Shijian} \sur{Li}}\email{shijianli@zju.edu.cn}
\author[1,2]{\fnm{Gang} \sur{Pan}}\email{gpan@zju.edu.cn}

\affil*[1]{\orgdiv{College of Computer Science and Technology}, \orgname{Zhejiang University}, \orgaddress{\street{38 Zheda Road}, \city{Hangzhou}, \postcode{310012}, \state{Zhejiang Province}, \country{China}}}

\affil[2]{\orgdiv{State Key Laboratory of Brain-Machine Intelligence}, \orgname{Zhejiang University}, \orgaddress{\street{38 Zheda Road}, \city{Hangzhou}, \postcode{310012}, \state{Zhejiang Province}, \country{China}}}

% \affil[3]{\orgdiv{Department}, \orgname{Organization}, \orgaddress{\street{Street}, \city{City}, \postcode{610101}, \state{State}, \country{Country}}}


\abstract{
The critical role of division of labor (DOL) in enhancing cooperation is well-recognized in real-world applications.
Consequently, many cooperative multi-agent reinforcement learning (MARL) methods have incorporated DOL mechanisms to improve cooperation among agents.
However, the lack of benchmark tasks specifically designed to evaluate and promote DOL and cooperation has limited the effective development and deployment of such mechanisms in cooperative MARL.
This gap between current cooperative MARL methods and practical applications underscores the need for evaluation tasks that explicitly require DOL and cooperation.
To address this gap, we propose the \textbf{C}omposite \textbf{T}asks \textbf{C}hallenge (\textbf{CTC}) — a suite of tasks explicitly designed to require both DOL and cooperation for successful task completion.
The CTC tasks are constructed based on two core design principles:
1) DOL is a necessary condition for task success;
2) Failure in any atomic subtask results in failure of the overall task.
The first principle emphasizes the necessity of DOL, while the second enforces the importance of cooperation, making both components essential for success in CTC tasks.
We evaluate nine representative cooperative MARL methods on the proposed CTC tasks.
Experimental results show that all methods consistently achieve zero test winning rates across all CTC tasks, highlighting the challenge of CTC tasks and the limitations of current methods.
To facilitate future research, we also introduce a guiding solution that achieves non-zero test winning rates on all tasks, thereby demonstrating the solvability of the CTC tasks.
However, the performance of this guiding solution remains suboptimal, further underscoring the value of CTC tasks as a challenging and meaningful testbed for advancing cooperative MARL research.
The CTC files and source code are available on https://github.com/Yurui-Li/CTC/.
}




\maketitle

\section{Introduction}\label{sec1}
Division of labor (DOL) is a fundamental organizing principle that enhances efficiency across both the biological world and human society.
In the biological world, DOL manifests in various forms, such as the specialized castes in social insects~\cite{holldobler0}, cellular differentiation in multicellular organisms, and the functional specialization observed in colonial marine invertebrates~\cite{dunn2006evolution}, and even bacteria~\cite{crespi2001evolution}.
In human society, DOL has been a cornerstone of economic and technological progress since Adam Smith’s seminal work "The Wealth of Nations"~\cite{smith2002inquiry}, and it is still ubiquitous in modern production systems.

Given its pervasive role in fostering cooperation and efficiency, DOL presents a natural paradigm for advancing cooperative policy learning in multi-agent reinforcement learning (MARL).
Accordingly, many MARL methods have incorporated DOL into their algorithmic frameworks, typically through one of three main paradigms: policy diversity, agent grouping, and hierarchical MARL.
Policy diversity~\cite{jiang2021emergence,mahajan2019maven,li2021celebrating} seeks to address the convergence to similar behaviors often observed in parameter-sharing architectures by encouraging agents to adopt distinct behavioral policies.
This implicit specialization leads to emergent DOL, where agents develop complementary behaviors tailored to different subtasks.
Agent grouping formalizes DOL by partitioning agents into functionally distinct subgroups.
Group membership may be determined based on predefined roles~\cite{wang2020roma}, sub-goals~\cite{christianos2021scaling}, task structures~\cite{yang2022ldsa}, or intrinsic agent capabilities~\cite{christianos2021scaling}.
Agents within a group typically share a policy, while differentiation between groups enables specialized execution across subtasks.
Hierarchical MARL architectures implement DOL via multi-level task decomposition.
The high-level controller manages task decomposition and subtask assignment, while the low-level controllers focus on executing these subtasks.
Such architectures align with DOL principles by facilitating systematic subtask specialization, whether through joint action space partitioning~\cite{wang2020rode}, learned subtask representations~\cite{yang2022ldsa}, or classification-based task selectors~\cite{li2024coordinating}.
Despite the theoretical integration of DOL in these methods, their practical effectiveness is constrained by a critical shortcoming: the lack of testbeds that explicitly require and reward DOL.

Existing MARL testbeds often only implicitly involve DOL, suffering from two key limitations: 1) DOL is not strictly necessary to complete the task; 2) failure in a single subtask does not directly cause task failure.
As a result, agents can succeed without completing all subtasks, undermining the evaluation and optimization of the policies aiming for DOL and cooperation.
For example, in the Multi-Agent Particle Environment (MPE)~\cite{lowe2017multi,mordatch2017emergence}, while tasks such as \textbf{cooperative communication} and \textbf{cooperative navigation} involve DOL, the former is limited by predefined roles, and the latter does not penalize partial subtask failure.
In Level-Based Foraging (LBF)~\cite{christianos2020shared,papoudakis2021benchmarking}, both DOL and non-DOL policies are viable, rendering DOL optional.
Similarly, in multi-robot warehouse environments (RWARE)~\cite{papoudakis2021benchmarking}, DOL is helpful but not critical, as the successful subtasks can compensate for the failed subtasks.
Tasks in StarCraft Multi-Agent Challenge (SMAC)~\cite{samvelyan2019starcraft}, SMACv2~\cite{ellis2022smacv2}, and GRF~\cite{kurach2020google} also allow task success through shared policies like focus firing or solo ball control, thus minimizing the need for explicit DOL.
Existing testbeds are insufficient for benchmarking and advancing DOL in cooperative MARL, as they do not explicitly require or evaluate DOL mechanisms.

To advance research of DOL and cooperation in cooperative MARL, we propose the \textbf{C}omposite \textbf{T}asks \textbf{C}hallenge (\textbf{CTC})—a suite of benchmark tasks designed to explicitly require and reward DOL to form successful cooperation.
We provide two design principles to ensure that CTC tasks have clear requirements for DOL:
1) DOL is a necessary condition for task success.
2) Failure in any atomic subtask results in failure of the overall task.
The first principle emphasizes the necessity of DOL, while the second enforces the importance of cooperation, making both components essential for success in CTC tasks.
Considering the diversity and complexity of real-world scenarios, we also incorporate \textbf{asymmetry} and \textbf{heterogeneity} to increase the realism of CTC tasks.
These designs make the CTC tasks more representative of real-world scenarios and highlight the need for sophisticated DOL mechanisms to form successful cooperation.
To evaluate existing methods under the CTC tasks, we select 8 representative cooperative MARL methods spanning the three paradigms (policy diversity, agent grouping, and hierarchical MARL) and a classic methods QMIX~\cite{rashid2018qmix,hu2021rethinking} as baselines.
Experimental results show that all baselines fail to solve the CTC tasks, consistently achieving zero test winning rates. This finding reveals a substantial gap between the theoretical capabilities of current methods and their practical performance, and raises concerns regarding the solvability of CTC tasks.
To address this issue, we introduce a guiding solution consisting of a rule-based external reward (RER) and an extended QMIX architecture (e-QMIX).
With the support of RER, e-QMIX achieves non-zero test winning rates across all CTC tasks, thereby demonstrating the solvability of CTC tasks.
Moreover, this guiding solution provides a useful reference for future research. 
However, RER is specifically tailored to the CTC tasks implemented in a particular environment and may not generalize to other settings, and its overall effectiveness remains limited.
These findings highlight both the inherent difficulty of the CTC tasks and their value as a benchmark for evaluating the ability of MARL methods to learn effective and reasonable DOL policies for cooperation. 
As such, CTC tasks have strong potential to drive the development of more advanced and practical cooperative MARL methods.

Our contributions can be summarized as follows:
\begin{itemize}
    \item Testbed: We introduce the CTC tasks, a novel testbed comprising solvable and practical tasks with explicit requirements for DOL, aimed at bridging the gap between MARL methods and real-world applications.
    \item Benchmark: We evaluate 9 representative cooperative MARL methods on the CTC tasks, demonstrating their current limitations in effectively leveraging DOL and cooperation.
    \item Guiding Solution: We propose a guiding solution to facilitate progress on CTC tasks, providing evidence that the tasks are both challenging and solvable, thus serving as a valuable guideline for future research.
\end{itemize}

\begin{figure}[t]
\begin{center}
\begin{tabular}{@{\extracolsep{\fill}}c@{}c@{}c@{}c@{\extracolsep{\fill}}}
            \includegraphics[scale=0.3]{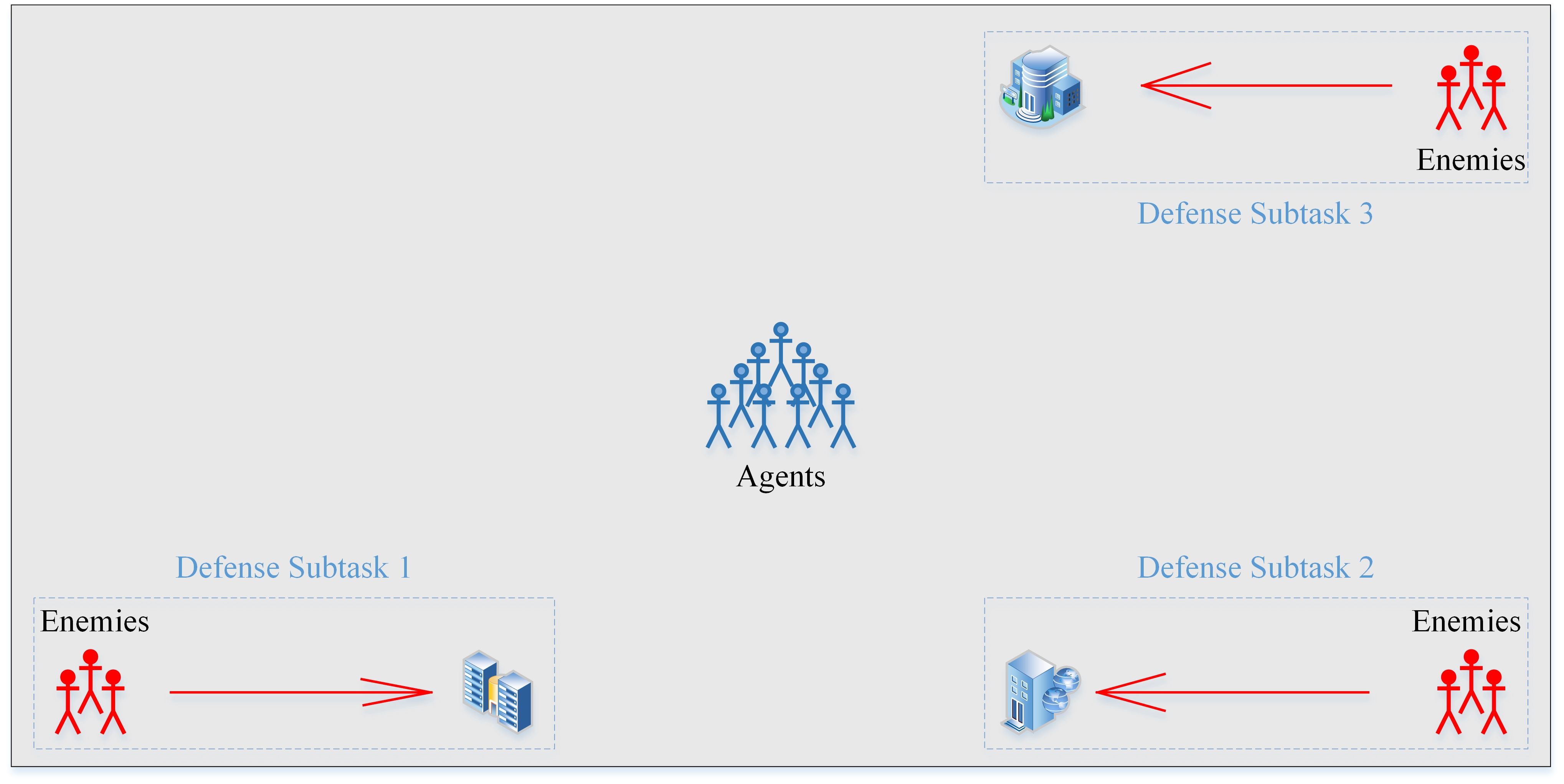} &
            \includegraphics[scale=0.3]{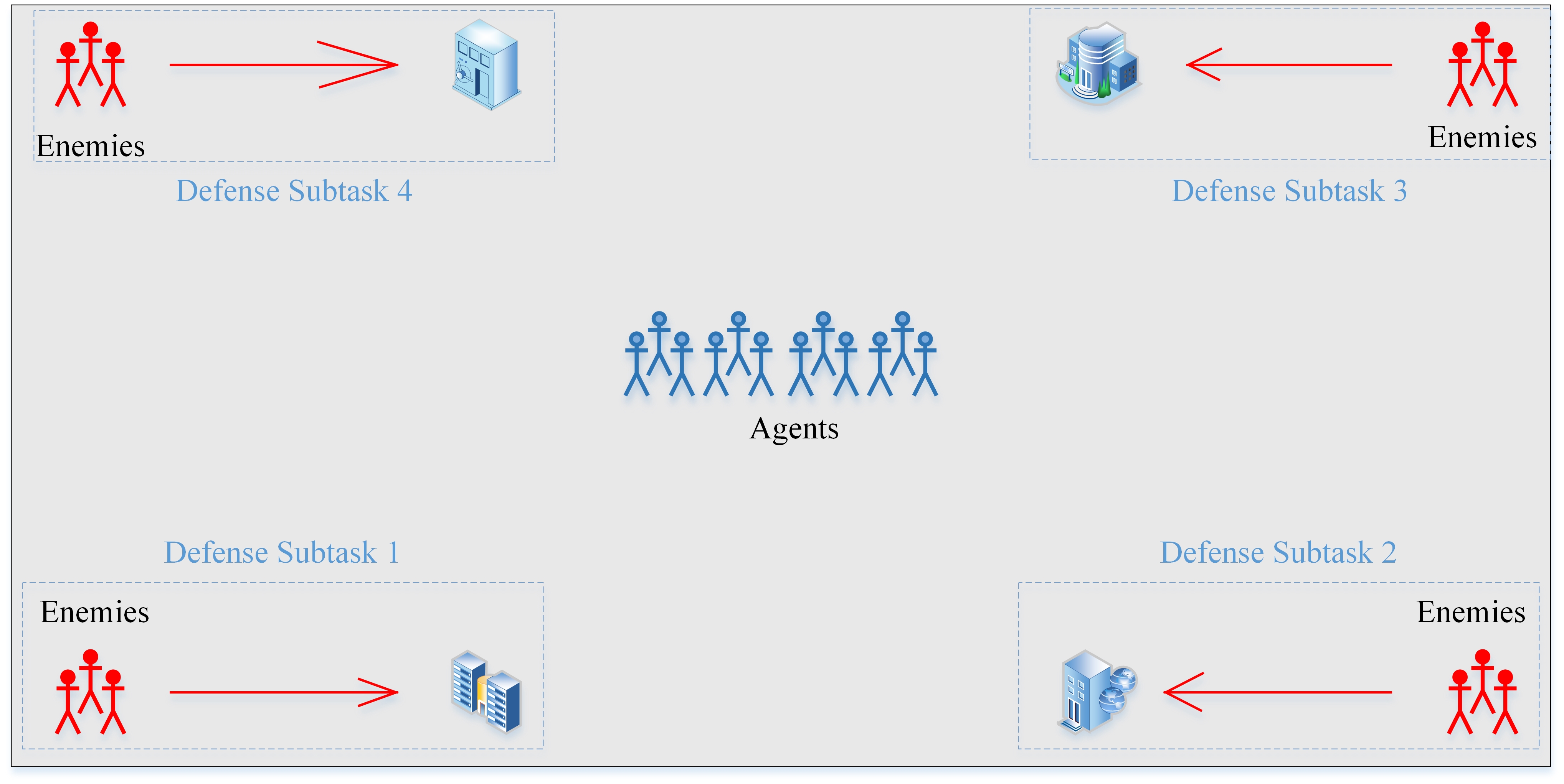} \\
             (a) Defense\_3\_Subtask (D3S)& (b) Defense\_4\_Subtask (D4S) \\
            \includegraphics[scale=0.3]{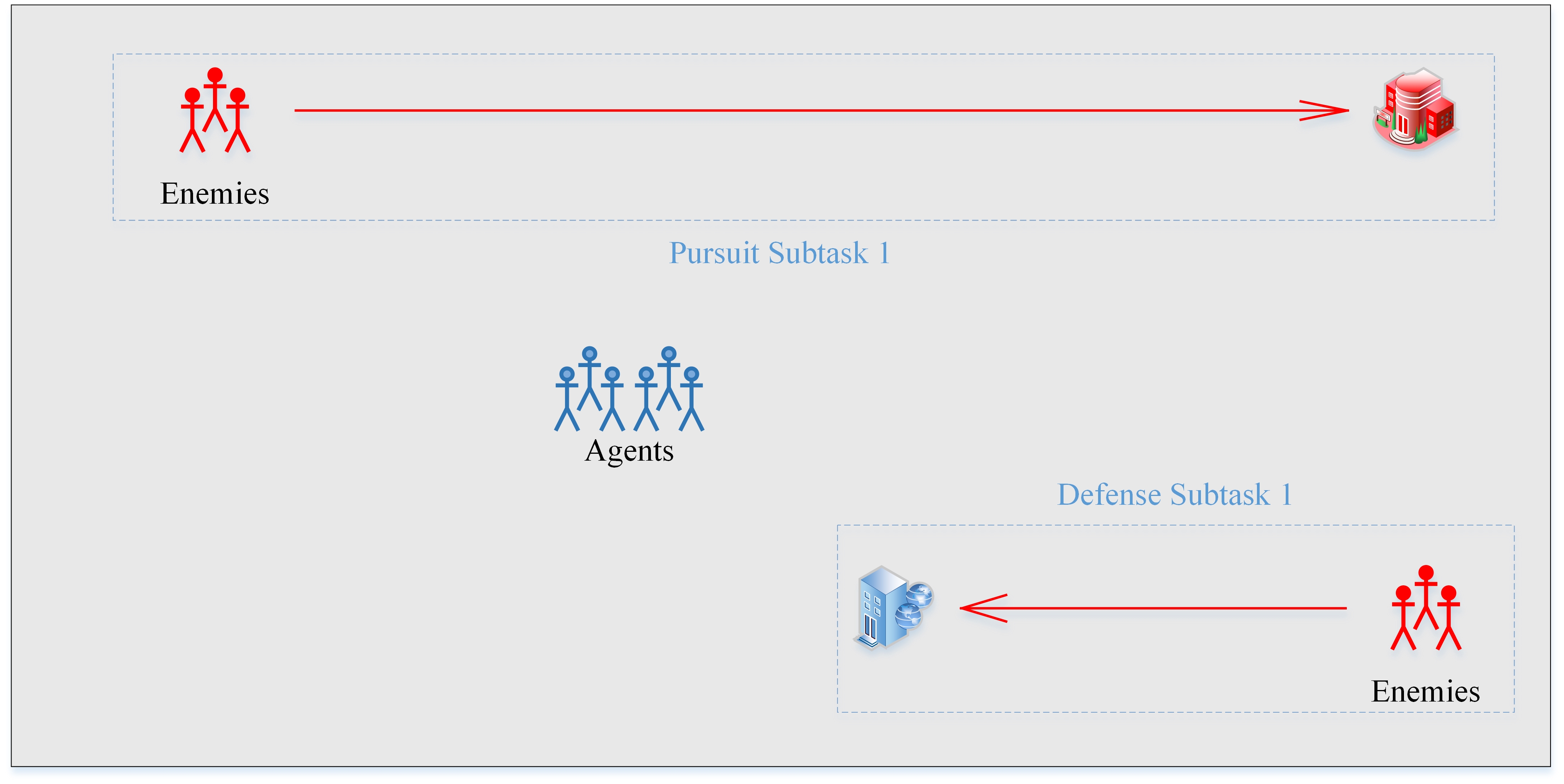} &
            \includegraphics[scale=0.3]{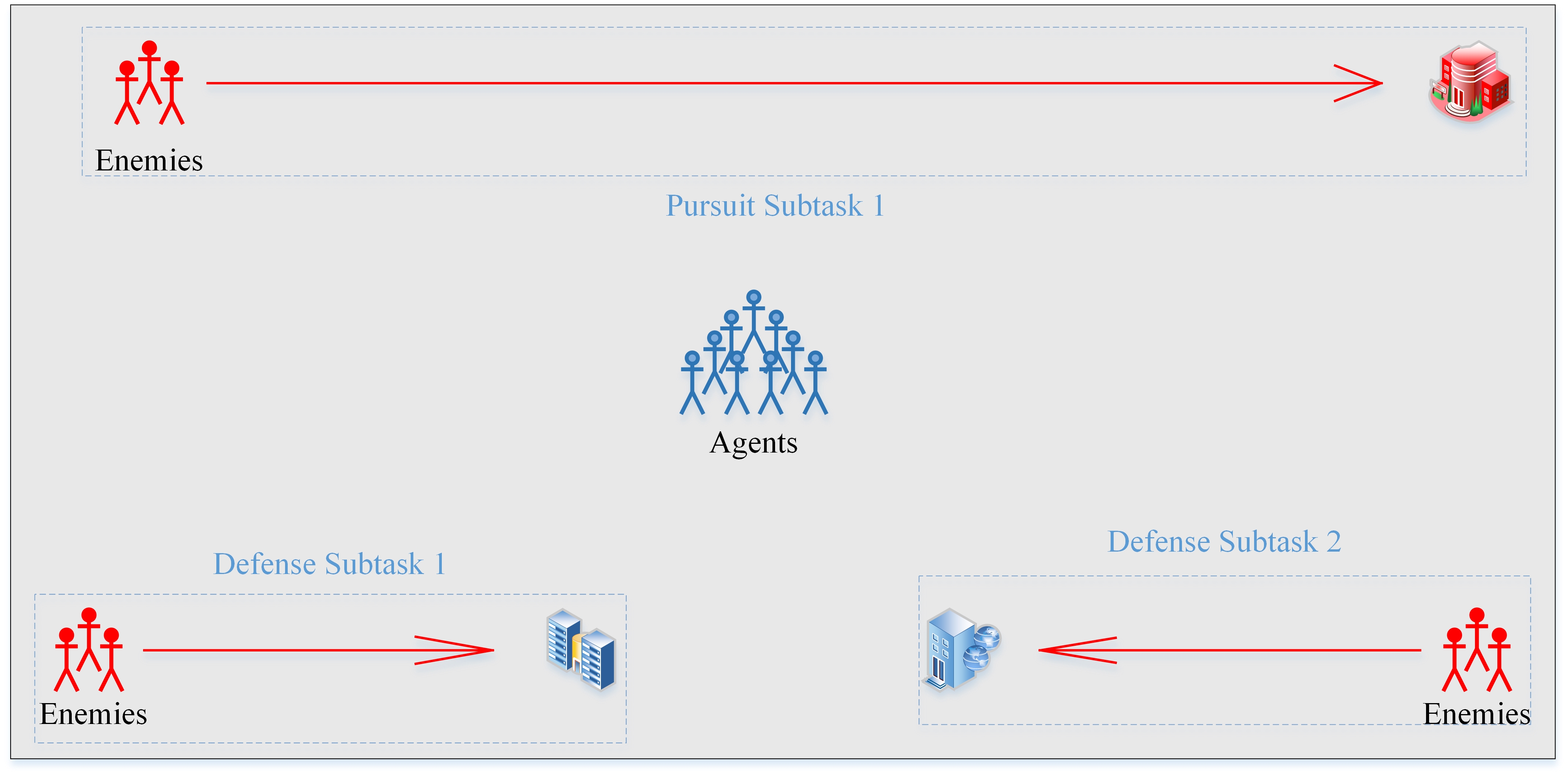} \\
               (c) Mixed\_2\_Subtask (M2S)& (d) Mixed\_3\_Subtask (M3S)\\
\end{tabular}
\caption{
The proposed CTC tasks are illustrated in the figure. 
Humanoid icons represent active entities, such as agents and enemies, while building icons denote base buildings. 
Agents and enemies are distinguished by color: blue icons represent agents, and red icons represent enemies. 
The figure also shows the initial positions of all agents and enemy units. 
The attack directions of enemy units are indicated by red arrows.
In each task, the number of agents and enemies is equal. 
Enemy units are divided into multiple groups, each assigned a specific objective, such as attacking an agent base building or retreating to their own base building. 
To complete the overall task, agents must allocate their forces appropriately across these objectives, ensuring that each enemy group is handled effectively. 
This setup requires agents to coordinate their behaviors and distribute their efforts across multiple simultaneous subtasks.
}
\label{fig:env}
\end{center}
\end{figure}

\section{Related Work}
\label{sec:a_rw}
In this section, we first demonstrate that the concept of division of labor (DOL) has been widely integrated into cooperative MARL methods.
We then discuss the limitations of the testbeds commonly used in current cooperative MARL research.

\subsection{Cooperative MARL Methods}
\label{sec:a_rw_cm}

We categorize cooperative MARL methods that incorporate DOL into three main paradigms: policy diversity, agent grouping, and hierarchical MARL.

\textbf{Policy diversity} promotes cooperation by enabling agents with a shared objective to learn distinct policies, thereby implicitly inducing specialization. For instance, EOI~\cite{jiang2021emergence} shows that emergent personalities can lead to differentiated roles, while CDS~\cite{li2021celebrating} explicitly encourages diversity via mutual-information regularization between agent identity and trajectories. Other works, such as DERE~\cite{jiang2022diverse} and SPD~\cite{jiang2022spd}, model inter-agent relationships or coordination structures to further facilitate role specialization. Collectively, these methods demonstrate that policy diversity is an effective mechanism for inducing DOL.

\textbf{Agent grouping} achieves DOL by partitioning agents into subgroups that specialize in different functions. Methods vary in how groups are formed: SEPS~\cite{christianos2021scaling} clusters agents based on trajectory embeddings, GACG~\cite{duan2024group} models cooperation dependencies and enforces inter-group diversity, and QTypeMix~\cite{fu2024qtypemix} leverages predefined agent types. Other approaches, including THGC~\cite{jiang2021multi}, VAST~\cite{phan2021vast}, and ROMA~\cite{wang2020roma}, group agents based on shared attributes or roles to enable coordinated learning. GoMARL~\cite{zang2024automatic} further introduces dynamic grouping mechanisms to improve scalability and efficiency. These methods highlight grouping as a structured way to realize DOL.

\textbf{Hierarchical MARL} implements DOL through multi-level task decomposition, where a high-level policy assigns subtasks and low-level policies execute them. Approaches such as RODE~\cite{wang2020rode} decompose tasks into role-based subtasks with shared policies, while LDSA~\cite{yang2022ldsa} and HSD~\cite{yang2019hierarchical} learn subtask representations or skills to support local coordination. DCC~\cite{li2024coordinating} formulates subtask assignment as a classification problem to guide agent behavior. These methods explicitly align DOL with task decomposition and specialization.

Despite incorporating DOL conceptually, these approaches are limited in practice by the lack of benchmark tasks that explicitly require and reward division of labor.

\subsection{Cooperative MARL Testbeds}
\label{sec:a_rw_ct}

We review commonly used cooperative MARL testbeds and analyze their relevance to DOL.

\textbf{Multi-Agent Particle Environment (MPE)}~\cite{lowe2017multi,mordatch2017emergence} consists of simple continuous-control tasks with discrete actions, dense rewards, and short horizons. Among its tasks, cooperative communication and cooperative navigation involve cooperation. In cooperative communication, a speaker and a listener have predefined roles, limiting the need for learned DOL. In cooperative navigation, agents must cover multiple landmarks while avoiding collisions, where DOL can improve efficiency. However, subtask success is not strictly tied to overall task success, making DOL beneficial but not necessary.

\textbf{Level-Based Foraging (LBF)}~\cite{christianos2020shared,papoudakis2021benchmarking} requires agents to collect items on a grid, with successful collection depending on the combined levels of participating agents. While agents may implicitly specialize (e.g., collecting items suited to their levels), such DOL is not required for task completion, as global success does not depend on subtask-level outcomes.

\textbf{Multi-Robot Warehouse (RWARE)}~\cite{papoudakis2021benchmarking,christianos2020shared} models partially observable warehouse logistics with sparse rewards. Agents must retrieve and deliver requested shelves through long action sequences. Although DOL can improve efficiency, it is not necessary, and subtask failures do not directly determine overall success.

\textbf{StarCraft Multi-Agent Challenge (SMAC/SMACv2)}~\cite{samvelyan19smac,ellis2022smacv2} provides combat-based tasks with partial observability and sparse rewards. SMACv2 introduces stochasticity and realism to mitigate overfitting. Effective strategies, such as focus fire or tactical retreat, primarily rely on coordinated but homogeneous behaviors rather than explicit role differentiation. Thus, DOL is not essential for optimal performance.

\textbf{Google Research Football (GRF)}~\cite{kurach2020google} is a complex, high-dimensional environment with realistic dynamics and sparse rewards. Although coordination is required, agents are homogeneous and fully capable of all actions. Consequently, DOL is not a necessary condition for success, as strong policies can emerge without explicit role specialization.

\textbf{Summary.} Across these benchmarks, while DOL can improve efficiency, it is generally not required for task success, and subtask outcomes rarely determine global performance. This limits their effectiveness for evaluating methods that explicitly aim to learn DOL.

\section{CTC Tasks}
\label{sec_ctc}
We design the CTC tasks around two core principles: (1) DOL must be a necessary condition for task completion, and (2) the failure of any individual atomic subtask results in overall task failure.
The first principle highlights the necessity of DOL, while the second underscores the importance of cooperation, thereby making both components essential for success in CTC tasks.
To satisfy \textbf{Principle 1}, each CTC task is constructed as a composition of multiple atomic subtasks, where successful completion of all subtasks is required to complete the CTC task.
This design also ensures the \textbf{decomposability} of CTC tasks, allowing them to be separated into disjoint atomic subtasks.
To satisfy \textbf{Principle 2}, we define task failure as occurring whenever any atomic subtask fails.
Furthermore, symmetry between subtasks and agent homogeneity is limited to reflecting simple real-world scenarios, resulting in limited simulation for practical applications.
To better capture the diversity and complexity of real-world cooperative tasks, we incorporate both \textbf{asymmetry} and \textbf{heterogeneity} into the design of tasks.
\textbf{Asymmetry} is introduced in two ways: by including multiple subtasks of the same type with varying configurations, and by combining subtasks of different types.
\textbf{Heterogeneity} is introduced through the deployment of agents with diverse types.
These design choices not only ensure that CTC tasks more accurately reflect real-world cooperative challenges but also reinforce the necessity of effective DOL and cooperation.
\textbf{Importantly, all atomic subtasks in a CTC task are initiated simultaneously and separated spatially, preventing any single agent from completing multiple subtasks at the same time.
Both agents and enemies are composed of the same types and quantities of units in all CTC tasks.
This ensures that performance differences arise from the challenges of the CTC tasks rather than imbalances in agent-enemy capabilities, allowing MARL methods to focus on learning effective DOL policies to form successful cooperation.}

\subsection{Atomic Subtasks}
Atomic subtasks serve as the foundational components of the CTC benchmark, facilitating the implementation of its two core design principles while also supporting asymmetry and heterogeneity in task structure.
To realize these objectives, we introduce two fundamental atomic subtasks: the defense subtask and the pursuit subtask.

In the defense subtask (bottom of Fig.~\ref{fig:env}(c)), a group of enemy units (red) attempts to occupy a designated base building (blue).
The distance between the enemy units’ starting point and the base building is set to 7.
The subtask is deemed a failure if any enemy unit successfully occupies the base building, while success is achieved if all enemy units are eliminated.
In the pursuit subtask (top of Fig.~\ref{fig:env}(c)), enemy units (red, left) attempt to retreat to their own base building (red, right), located 21 away from their starting position.
As in the defense subtask, failure occurs if any enemy unit reaches its base building, while success requires the complete elimination of all enemy units.

Although the pursuit and defense subtasks share structural similarities, they \textbf{differ substantially in enemy behavior}.
In the pursuit subtask, enemy units do not retaliate when attacked and instead persistently advance toward their base building.
By contrast, in the defense subtask, enemy units engage in combat when attacked.
This distinction effectively makes the pursuit subtask \textbf{higher priority}: agents must eliminate the retreating enemy units within 21 steps of subtask initiation to avoid failure.
In the defense subtask, enemy units occupy the base building only after defeating nearby agents within their attack range.
If no agents are within range, they capture the base building 7 steps after the subtask begins.
Thus, the pursuit subtask is inherently time-limited, whereas the defense subtask is contingent on combat outcomes.

\subsection{CTC Implementation}
\label{subsec_ctc_implement}
We implement the CTC tasks within SMAC~\cite{samvelyan19smac} for two primary reasons.
First, SMAC is the most widely adopted benchmark in cooperative MARL research.
By building on SMAC’s foundational settings while adhering to the design principles of CTC, we can focus on task construction without being constrained by low-level implementation details.
Second, leveraging SMAC ensures that CTC tasks can be seamlessly integrated into most existing MARL methods without requiring modifications to the original source code, thereby substantially reducing the implementation burden for researchers.
On this basis, we construct four CTC tasks.
Fig.~\ref {fig:env} illustrates the spatial configuration of each subtask.

We preserve all of SMAC’s original settings (details in \cite{samvelyan19smac}), with additional rules introduced solely to enforce the CTC design principles.
In all CTC tasks, task success requires the completion of all atomic subtasks, satisfying \textbf{Principle 1}, while task failure results from the failure of any atomic subtask, thereby fulfilling \textbf{Principle 2}.
The two tasks illustrated in Fig.~\ref{fig:env}(a–b) demonstrate \textbf{asymmetry} through the use of atomic subtasks of the same type but with different configurations.
Their configurations vary in the number and type of enemy units across atomic subtasks, introducing asymmetry in atomic subtasks within the same task.
In addition, the number of atomic subtasks in these two tasks increases progressively (from 3 to 4), thereby gradually increasing the complexity and difficulty of the tasks and establishing a smooth evaluation route.
In general, a greater number of atomic subtasks not only elevates task complexity but also more accurately reflects the demands of real-world cooperative scenarios. 
Methods capable of managing a wider range and larger number of subtasks demonstrate higher applicability and improved scalability, both in terms of task diversity and multi-agent cooperation.
The two tasks in Fig.~\ref{fig:env}(d–e) exemplify \textbf{asymmetry} through the composition of different types of atomic subtasks.
In practical applications, atomic subtasks are rarely homogeneous, and the degree of heterogeneity directly impacts task complexity and difficulty.
Tasks composed of similar subtasks are generally easier than those involving dissimilar ones.
Task Mixed\_2\_Subtask (Fig.~\ref{fig:env}(c)) includes one defense subtask and one pursuit subtask, representing a balanced composition of dissimilar atomic subtasks.
Task Mixed\_3\_Subtask (Fig.~\ref{fig:env}(d)) comprises one pursuit and two defense subtasks, introducing an imbalanced proportion of atomic subtask types.
These variations allow CTC to more accurately reflect the complexity and irregularity found in real-world cooperative scenarios.
To further introduce \textbf{heterogeneity}, we utilize the three Terran unit types from StarCraft II—Marine, Marauder, and Medivac—each with distinct capabilities.
Marine is light infantry with fast attack speed but relatively low health and damage.
Marauder, in contrast, has higher health and armor, and deals more damage but attacks more slowly.
Medivac is a non-combat unit with healing capabilities, crucial for sustaining teammates during engagements.
Table~\ref{tab:hea} demonstrates the number and types of agents and enemies of each CTC task.
The heterogeneous nature of the agents imposes additional demands on MARL methods, as successful cooperation must consider both task assignment and agent capabilities.
This design choice enhances the practical relevance of CTC tasks for evaluating cooperative MARL methods.

\begin{table}
  \caption{Agents and enemies setting of CTC tasks.}
  \label{tab:hea}
  \centering
  \begin{tabular}{lccccc}
    \toprule
    Tasks  & Agents & Subtask 1 & Subtask 2 & Subtask 3 & Subtask 4 \\
    \midrule
        Defense\_3\_Subtask  & \makecell{3 Marine \\ 3 Marauder \\ 3 Medivac}  & \makecell{1 Marauder \\ 1 Medivac} & \makecell{1 Marine \\ 1 Marauder \\ 1 Medivac} &\makecell{2 Marine \\ 1 Marauder \\ 1 Medivac} &/ \\
        \hline
        Defense\_4\_Subtask  & \makecell{4 Marine \\ 4 Marauder \\ 4 Medivac}  & \makecell{ 1 Marine \\ 1 Medivac} & \makecell{ 1 Marauder \\ 1 Medivac} &\makecell{1 Marine \\ 1 Marauder \\ 1 Medivac} &\makecell{2 Marine \\ 2 Marauder \\ 1 Medivac} \\
        \hline
        Mixed\_2\_Subtask  & \makecell{2 Marine \\ 2 Marauder \\ 2 Medivac}  & \makecell{1 Medivac} & \makecell{2 Marine \\ 2 Marauder \\ 1 Medivac} &/ &/ \\
        \hline
        Mixed\_3\_Subtask  & \makecell{3 Marine \\ 3 Marauder \\ 3 Medivac}  & \makecell{1 Medivac} & \makecell{1 Marine \\ 1 Marauder \\ 1 Medivac} &\makecell{2 Marine \\ 2 Marauder \\ 1 Medivac} &/ \\
    \bottomrule
  \end{tabular}
\end{table}

\begin{figure}[t]
\begin{center}
\begin{tabular}{@{\extracolsep{\fill}}c@{}c@{}c@{}c@{\extracolsep{\fill}}}
            \includegraphics[scale=0.23]{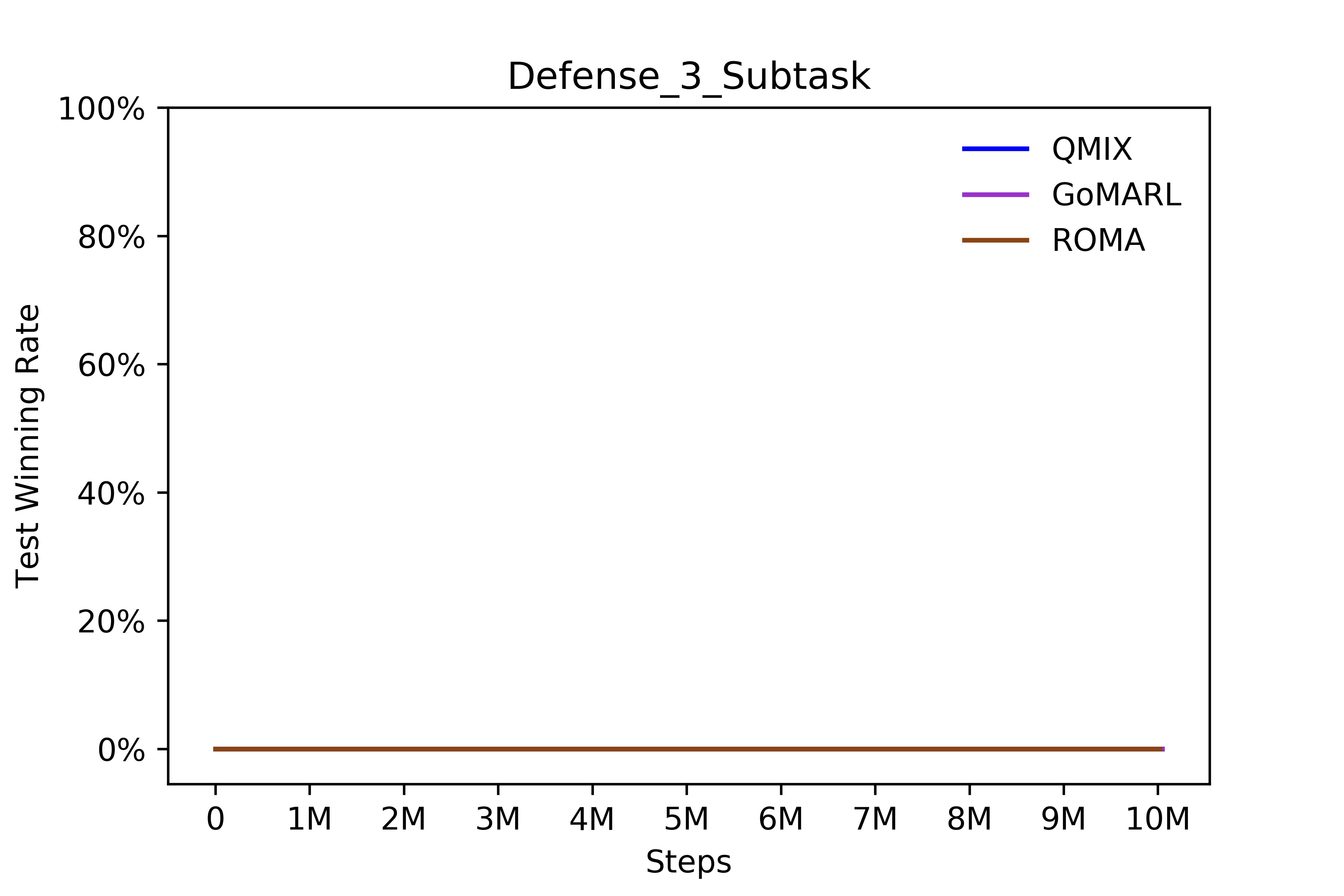} &
            \includegraphics[scale=0.23]{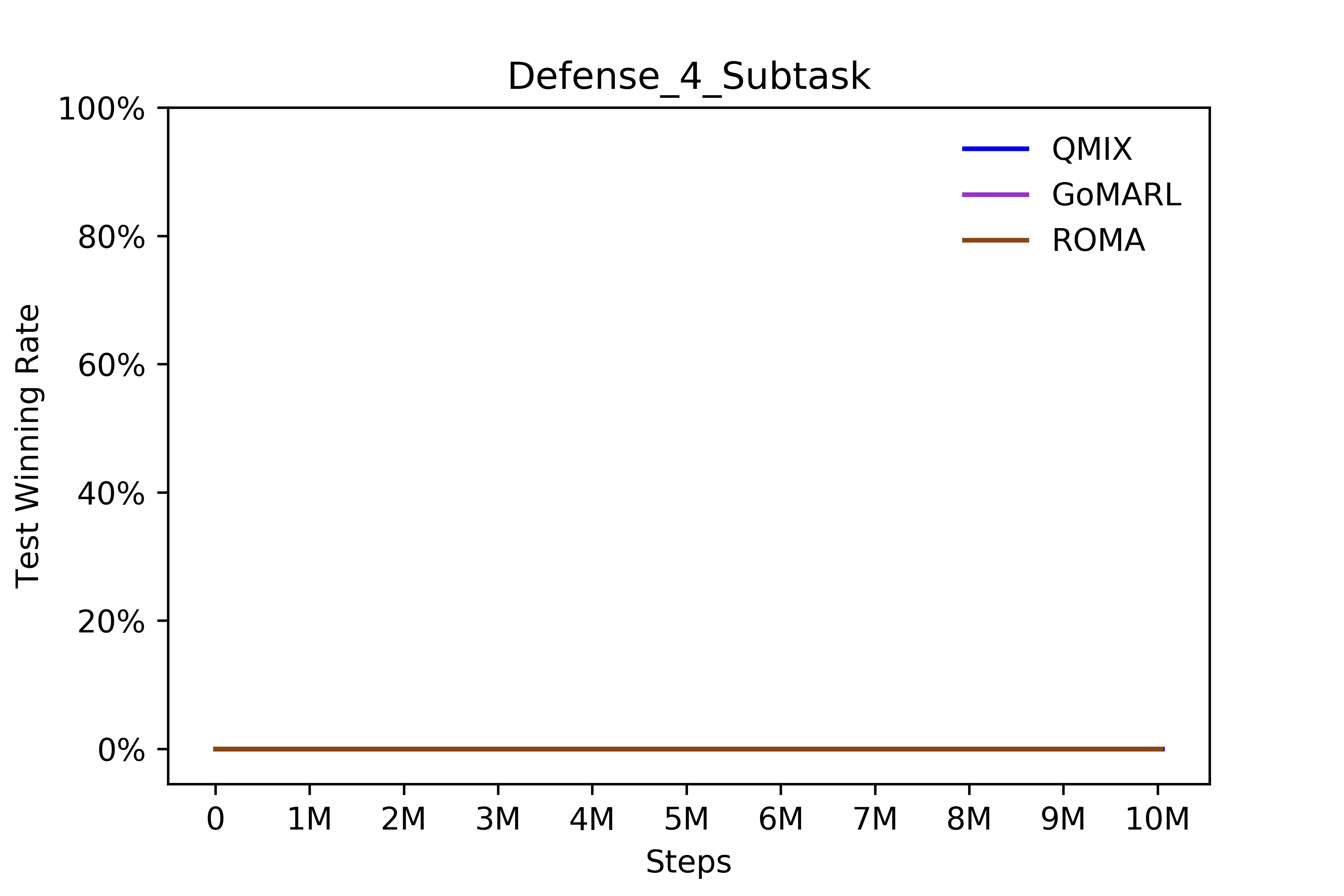} &
            \includegraphics[scale=0.23]{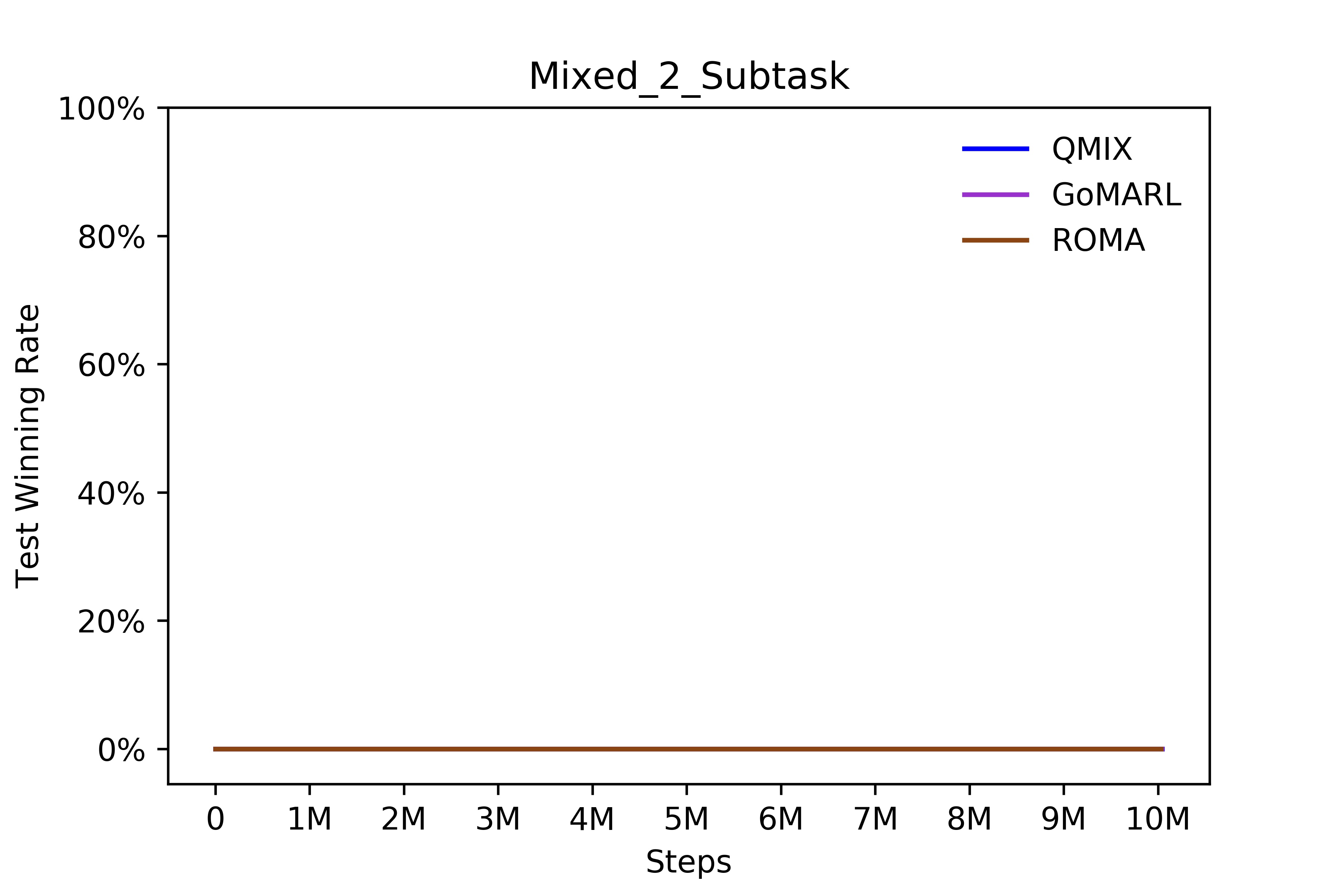} &
            \includegraphics[scale=0.23]{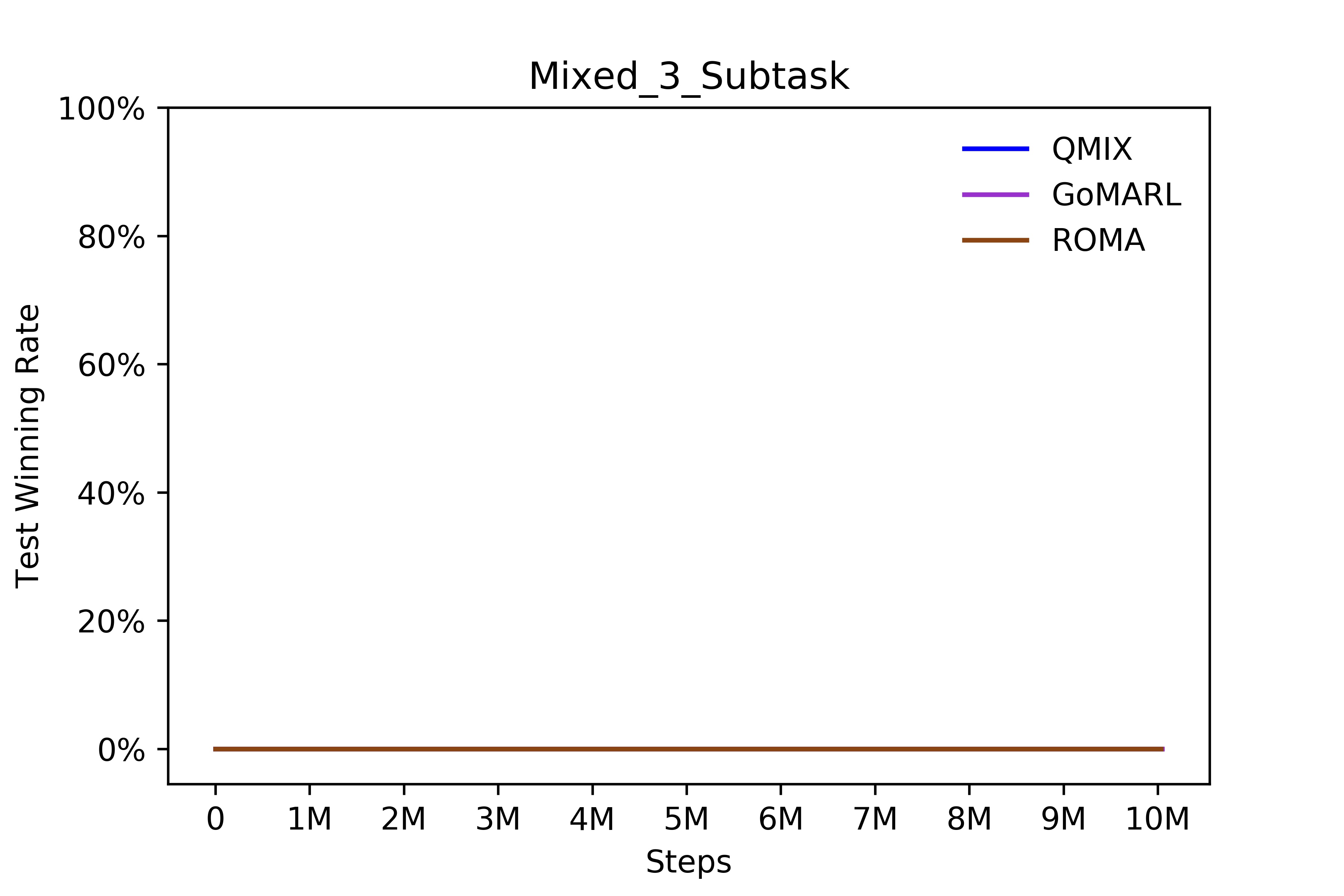}\\
            (a1) & (b1) & (c1) & (d1) \\
            \includegraphics[scale=0.23]{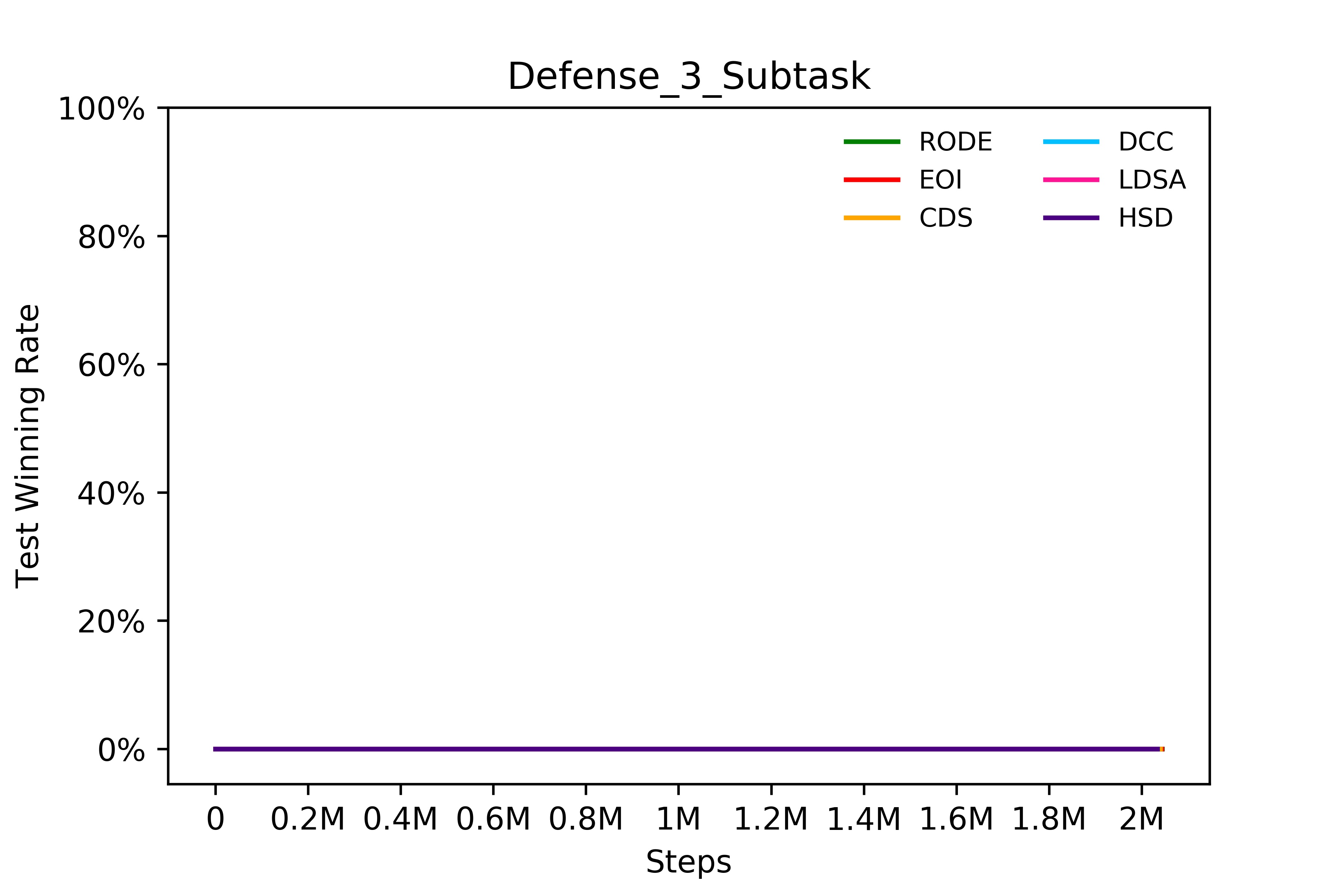} &
            \includegraphics[scale=0.23]{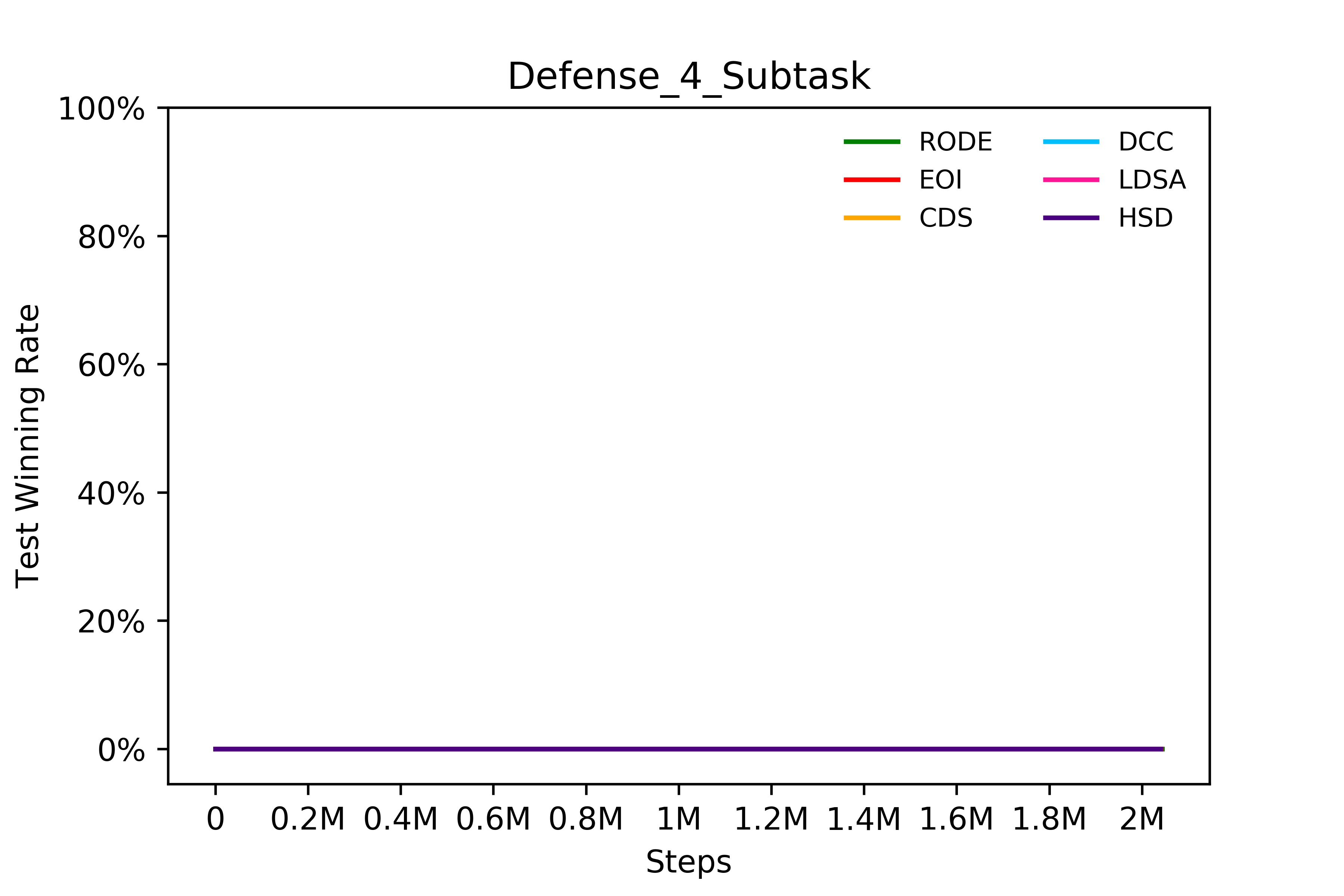} &
            \includegraphics[scale=0.23]{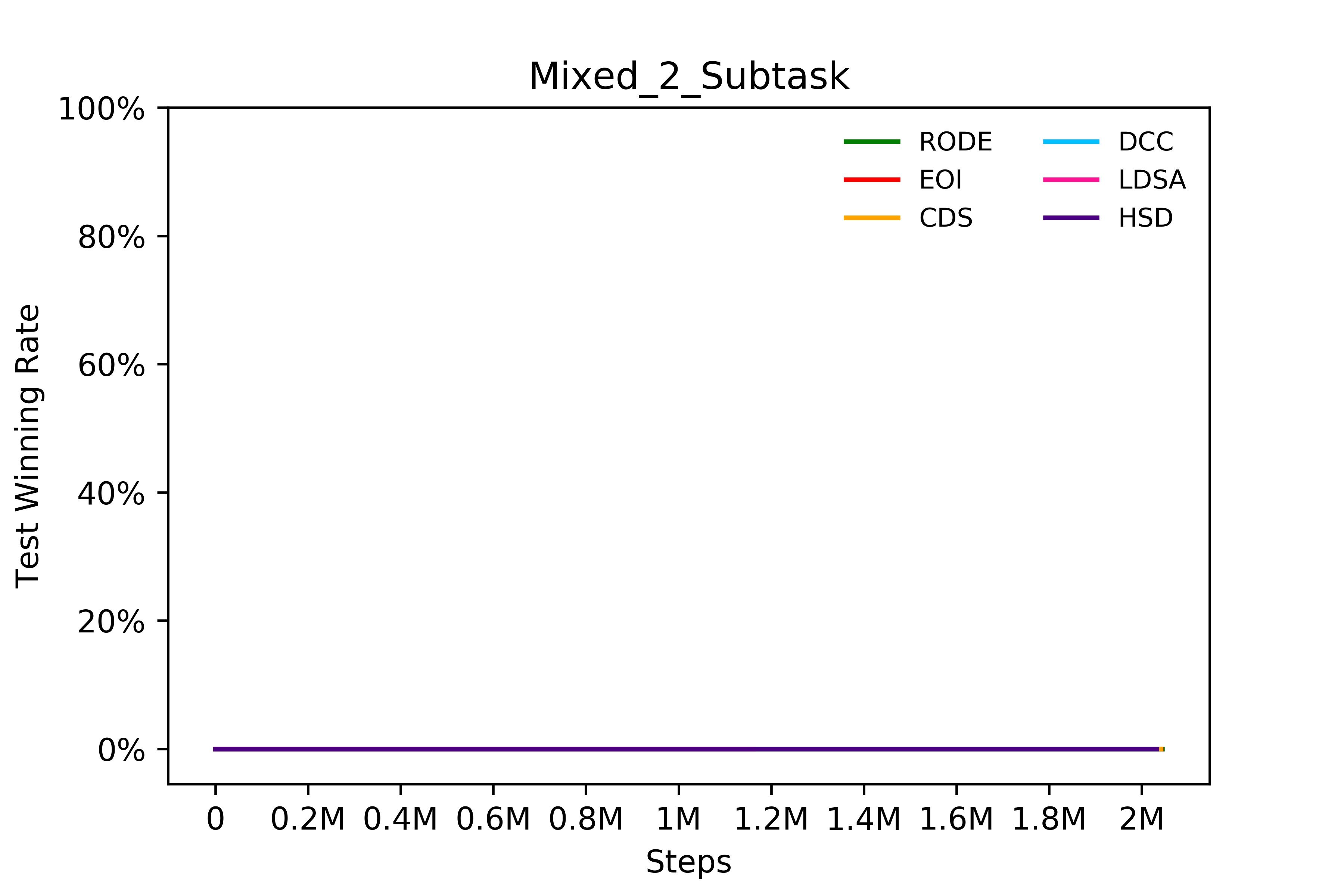} &
            \includegraphics[scale=0.23]{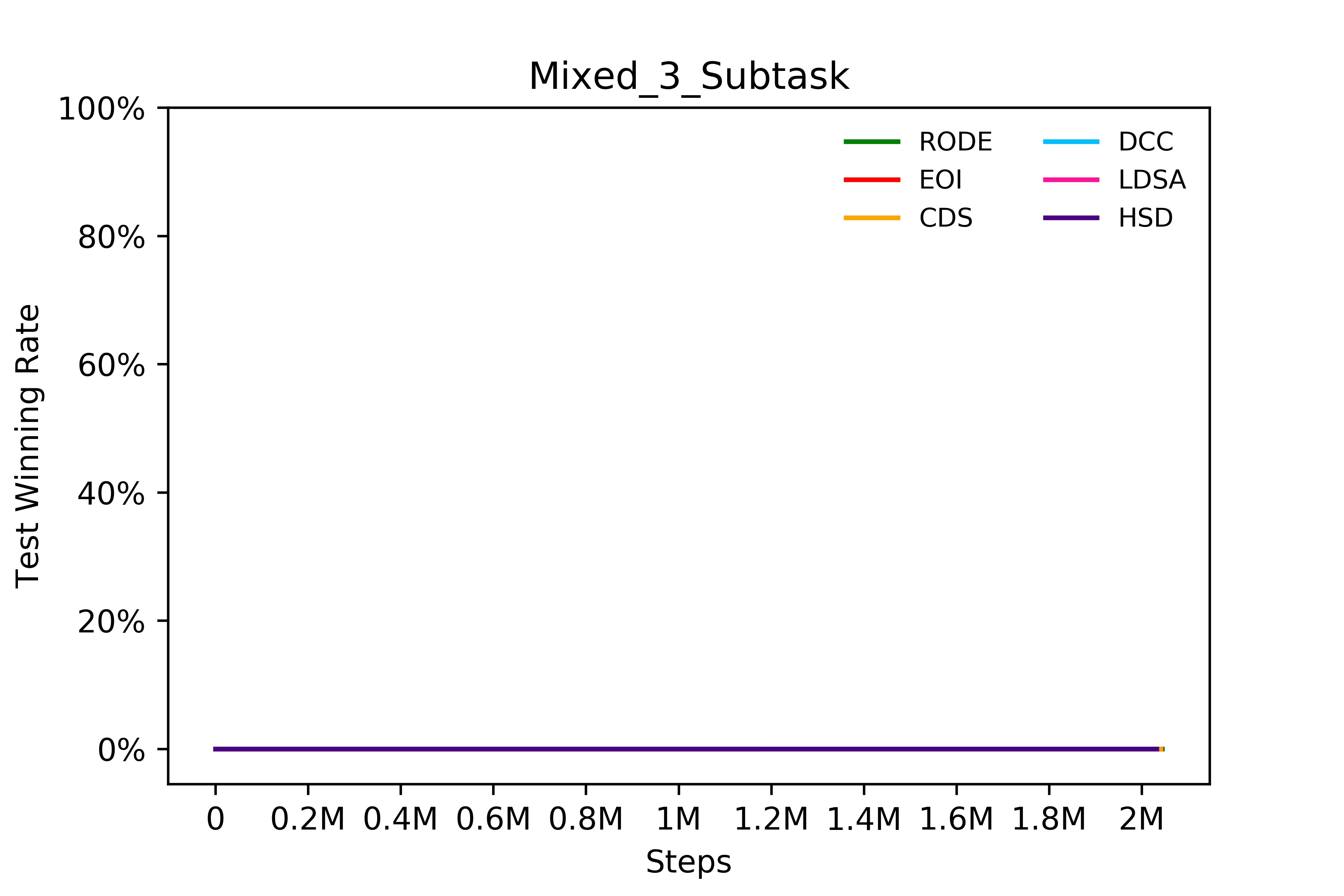}\\
            (a2) & (b2) & (c2) & (d2) \\
\end{tabular}
\caption{Performance of baselines on CTC tasks.}
\label{fig:performance}
\end{center}
\end{figure}

\begin{figure}
\begin{center}
\begin{tabular}{@{\extracolsep{\fill}}c@{}c@{}c@{}c@{\extracolsep{\fill}}}
            \includegraphics[scale=0.23]{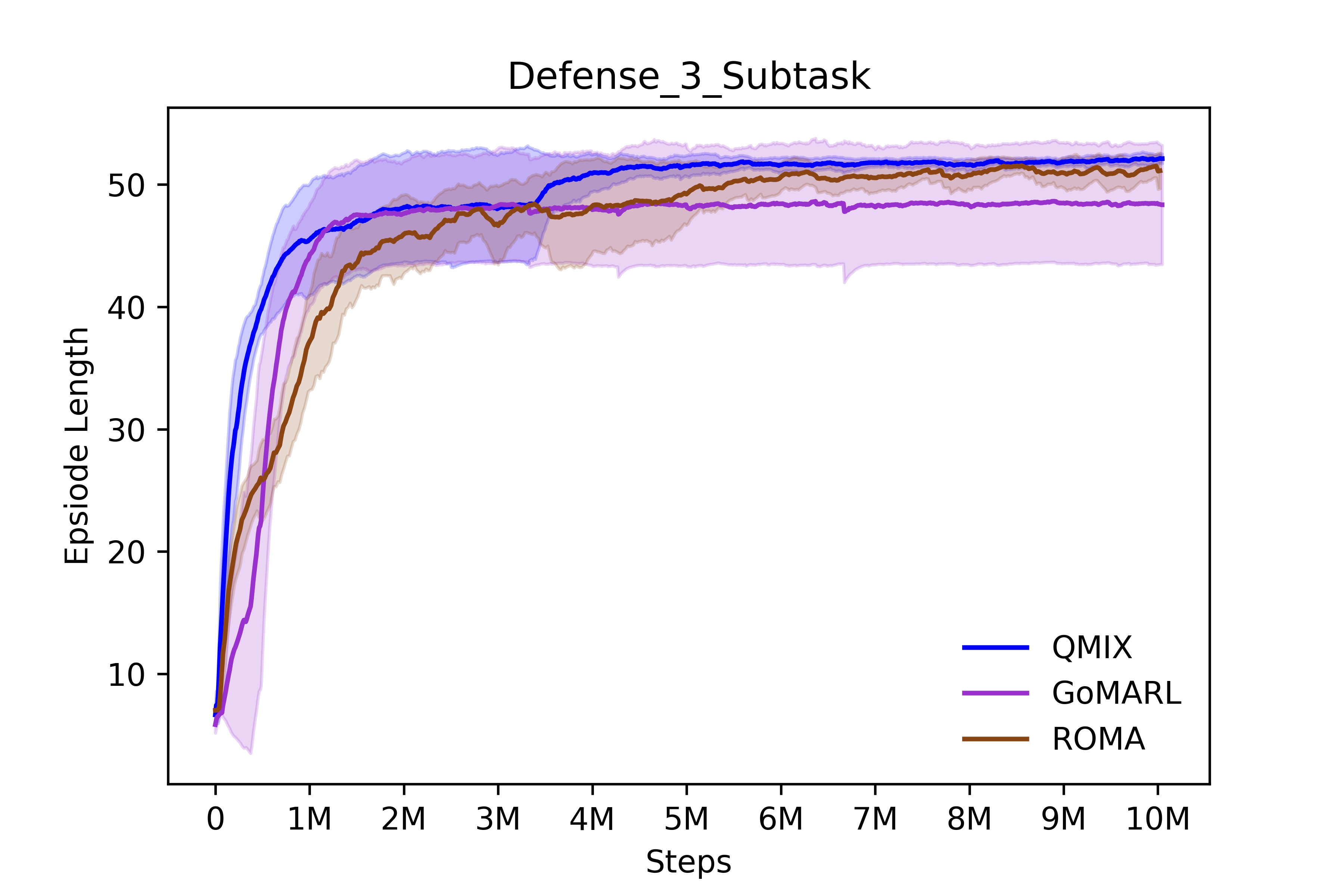} &
            \includegraphics[scale=0.23]{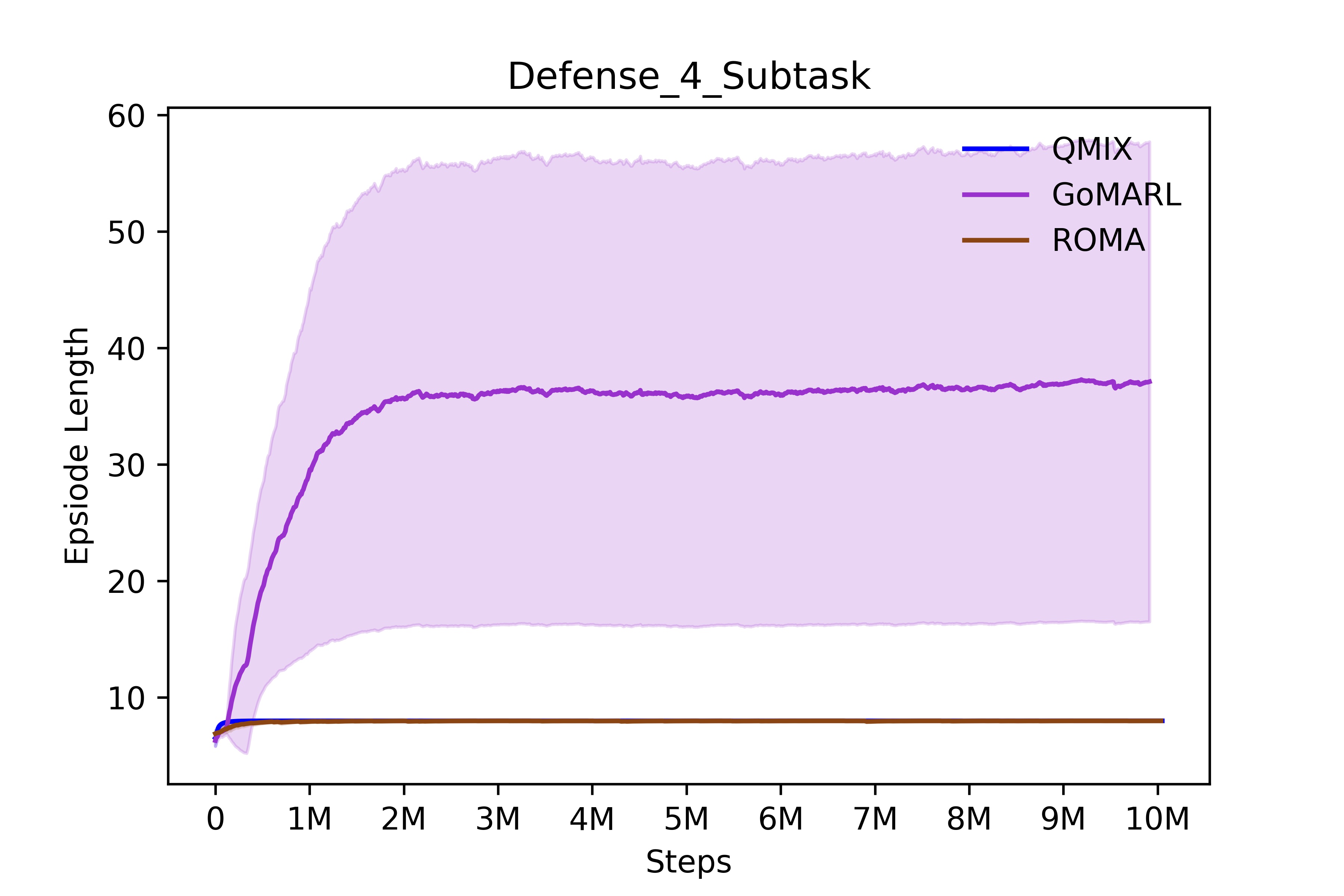} &
            \includegraphics[scale=0.23]{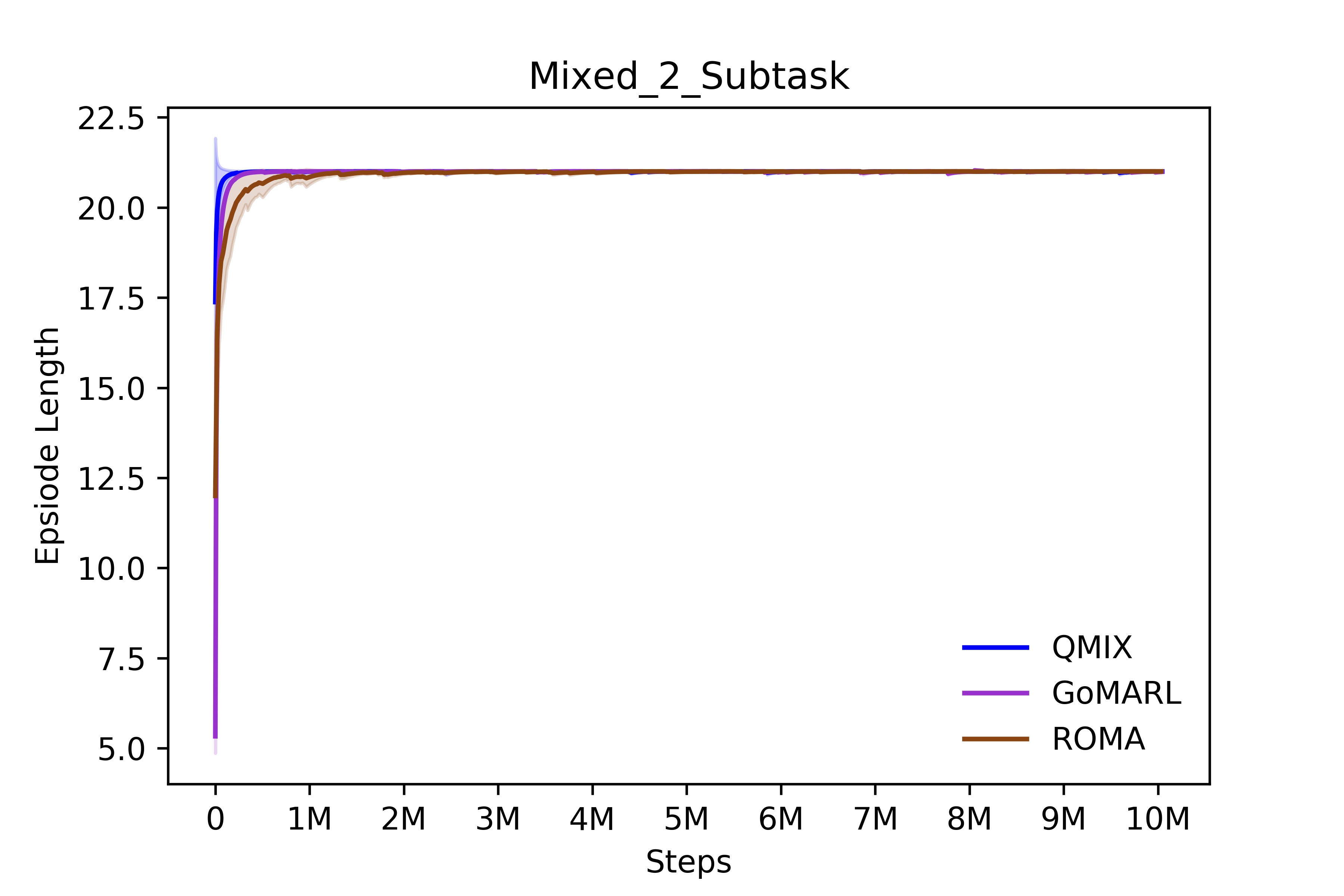} &
            \includegraphics[scale=0.23]{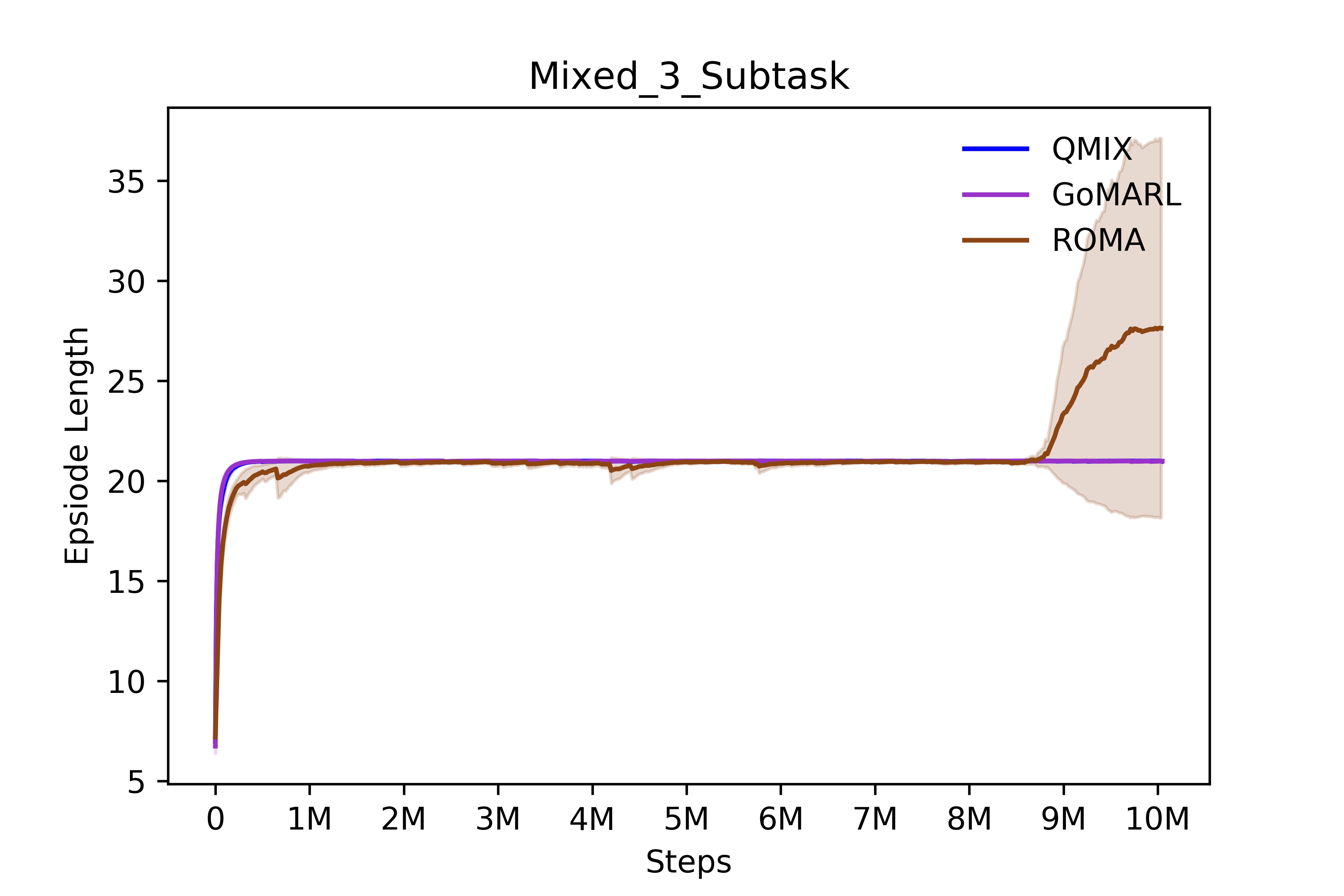}\\
            (a1) & (b1) & (c1) & (d1) \\
            \includegraphics[scale=0.23]{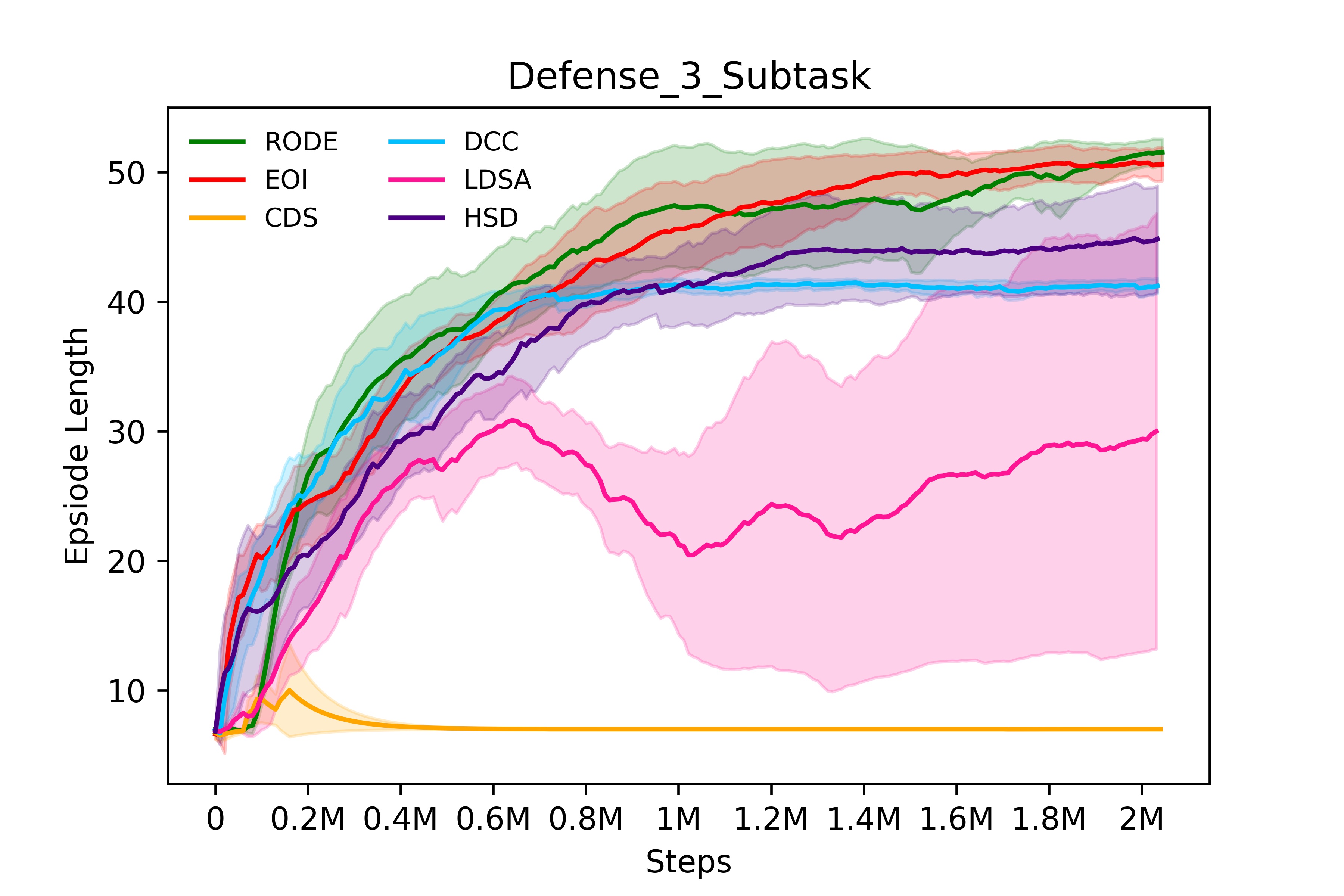} &
            \includegraphics[scale=0.23]{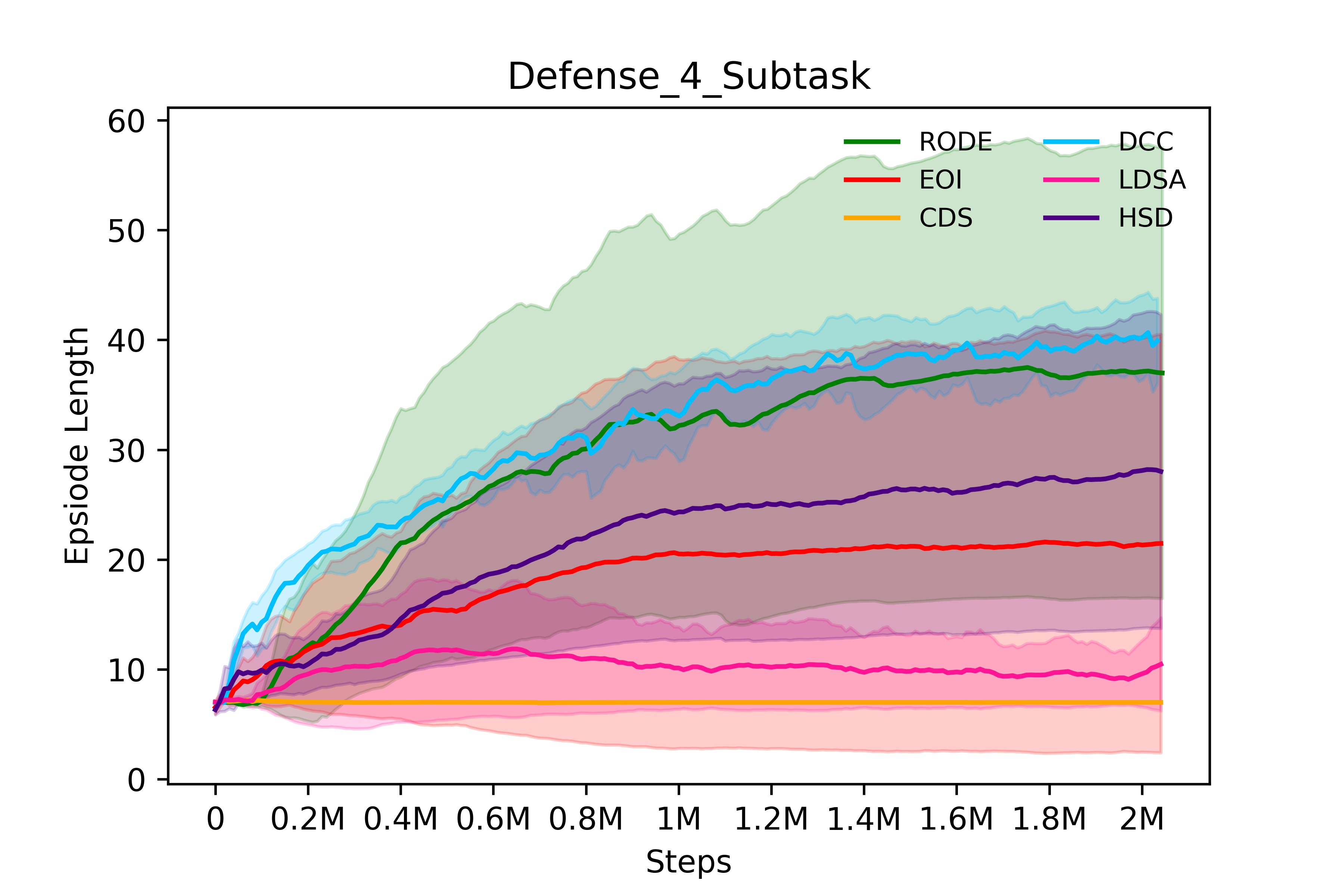} &
            \includegraphics[scale=0.23]{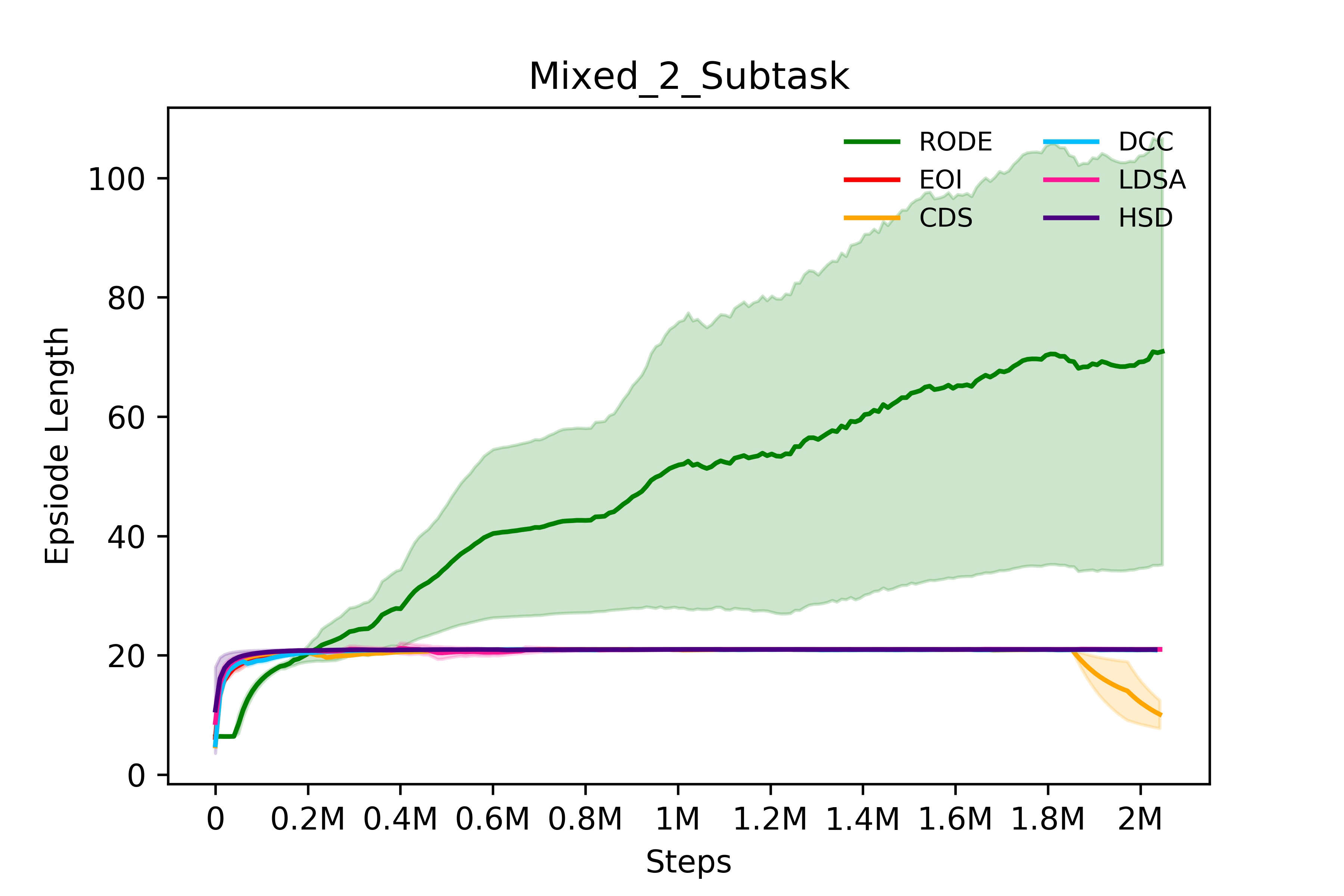} &
            \includegraphics[scale=0.23]{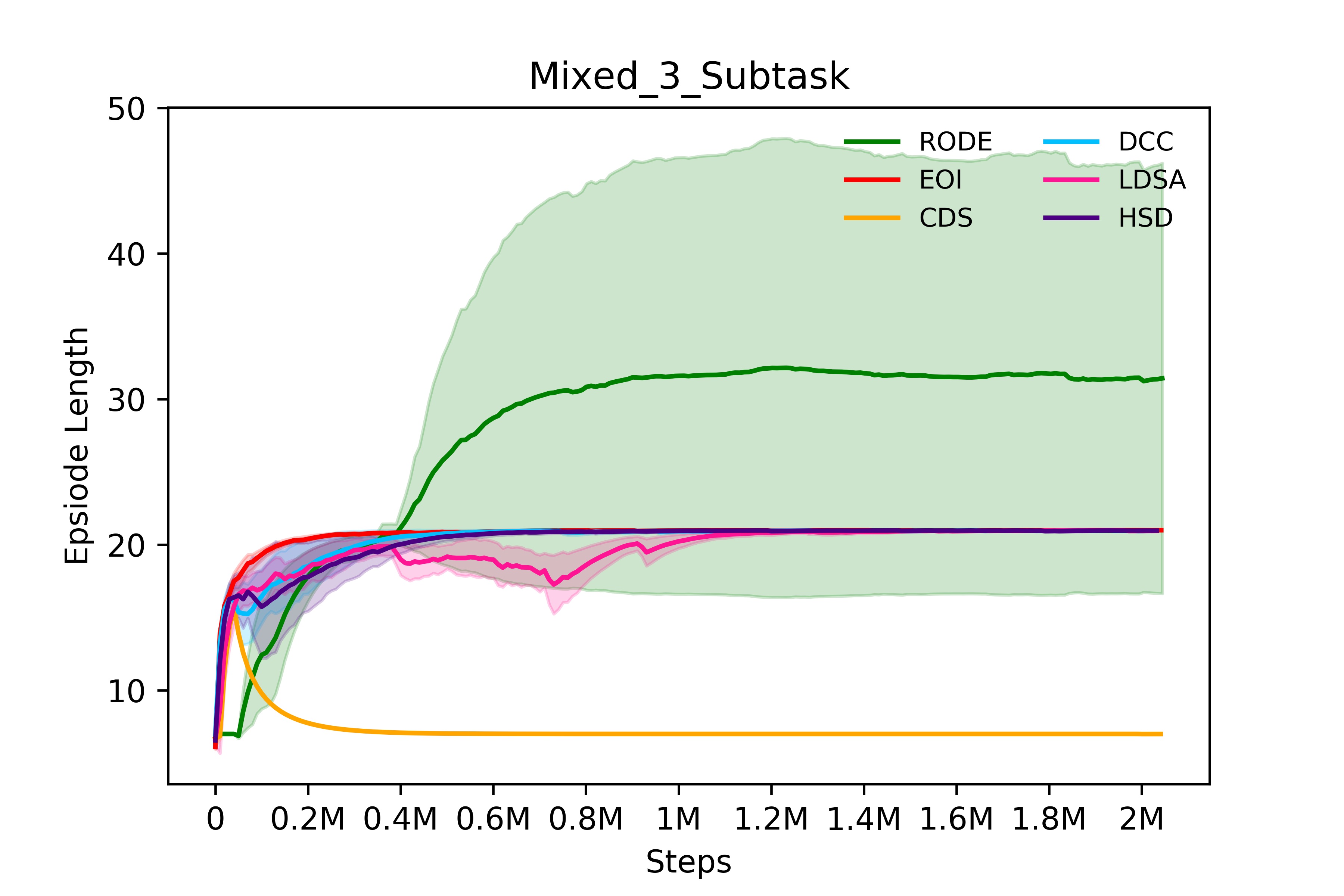}\\
            (a2) & (b2) & (c2) & (d2) \\
\end{tabular}
\caption{Episode length of baselines on CTC tasks.}
\label{fig:epLength}
\end{center}
\end{figure}

\section{Benchmark}
\label{sec_bench}
In this section, we present experimental evaluations to assess the performance of cooperative MARL methods on CTC tasks.
To evaluate the effectiveness of current state-of-the-art (SOTA) methods on CTC tasks, we select eight representative MARL methods spanning three key categories: policy diversity, agent grouping, and hierarchical MARL.
The selected methods are: EOI~\cite{jiang2021emergence}, DCC~\cite{li2024coordinating}, CDS~\cite{li2021celebrating}, RODE~\cite{wang2020rode}, ROMA~\cite{wang2020roma}, GoMARL~\cite{zang2024automatic},  LDSA~\cite{yang2022ldsa}, and HSD~\cite{yang2019hierarchical}.
Additionally, we include the classic cooperative MARL method QMIX~\cite{rashid2018qmix,hu2021rethinking} as a baseline.
We utilize the official implementations of each method and retain their best hyperparameter configurations to ensure fair comparisons.
Notably, all selected baselines are implemented using the PyMARL framework\footnote{https://github.com/oxwhirl/pymarl}, though they differ in execution strategies.
Specifically, QMIX, GoMARL, and ROMA employ parallel runners, executing multiple environments concurrently to collect data.
In contrast, the remaining methods utilize episode runners, operating one environment at a time.
Due to this divergence, the interpretation of the global step—used to track the number of environment interactions—differs across implementations.
To address this inconsistency, we categorize the methods into two groups based on runner type when presenting performance plots.
The primary evaluation metric is the test winning rate.
We conduct 32 evaluation episodes without exploration and report both the mean and standard deviation of the winning rates across five independent random seeds.

As shown in Fig.~\ref{fig:performance}, \textbf{all baselines achieve a test winning rate of zero across all CTC tasks}, rendering direct comparisons based on this metric uninformative.
Therefore, rather than comparing performance quantitatively, we focus on analyzing the underlying reasons for the consistent failure of these baselines in CTC tasks.

\subsection{Analysis}
\label{chap:basicPerformance}

Notably, episode length provides insight into how well a method handles specific atomic subtasks:
1) An episode length of less than 7 suggests that at least one defense subtask fails.
2) A length between 7 and 21 indicates that the pursuit subtask fails.
3) A length greater than 21 implies successful completion of the pursuit subtask, but it does not mean the defense subtask is completed.

\begin{figure}[t]
\begin{center}
\begin{tabular}{@{\extracolsep{\fill}}c@{}c@{\extracolsep{\fill}}}
            \includegraphics[scale=0.3]{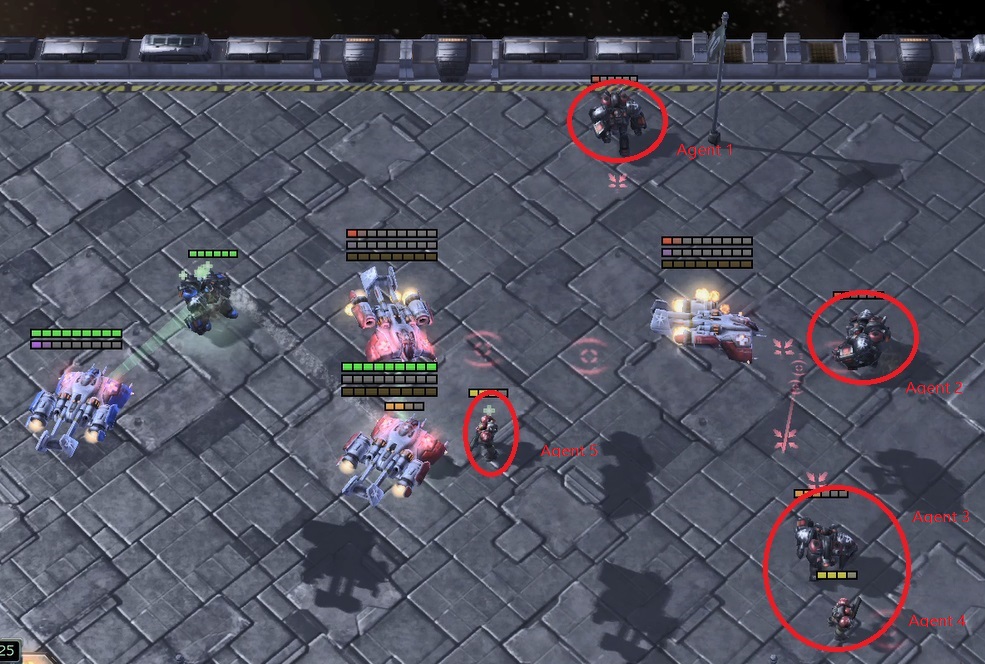} &
            \includegraphics[scale=0.25]{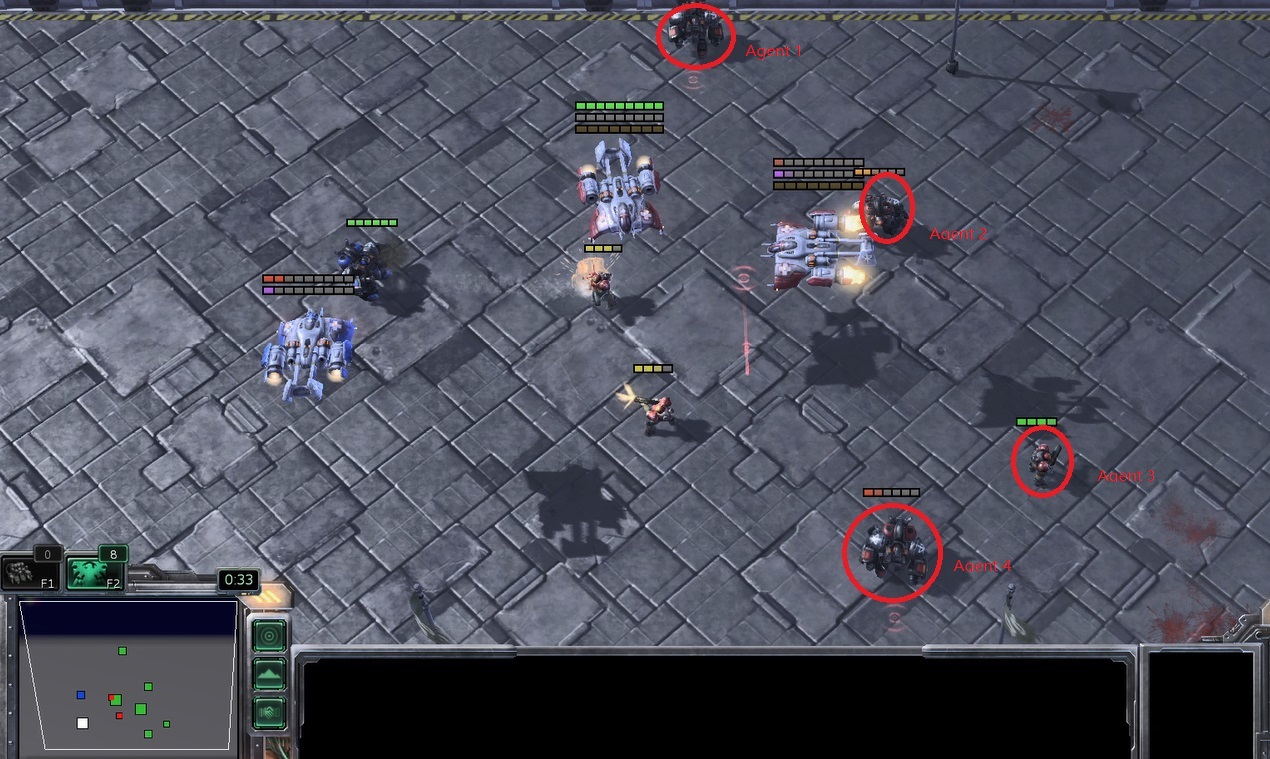} \\
            (a) & (b) \\
            \includegraphics[scale=0.4]{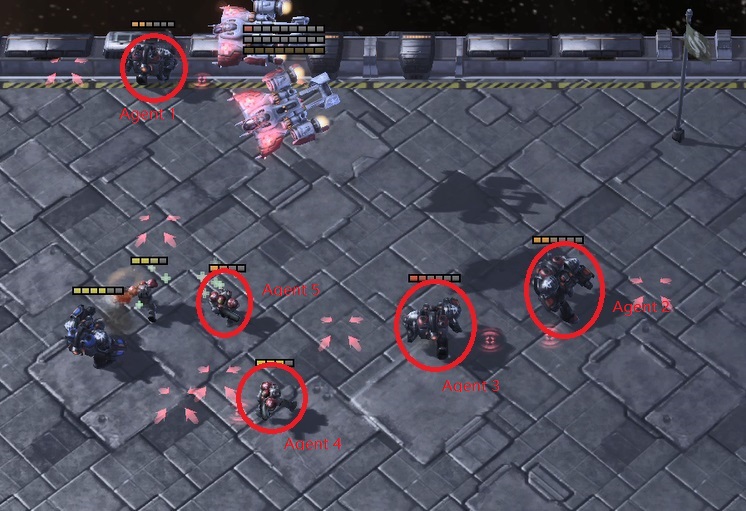} &
            \includegraphics[scale=0.3]{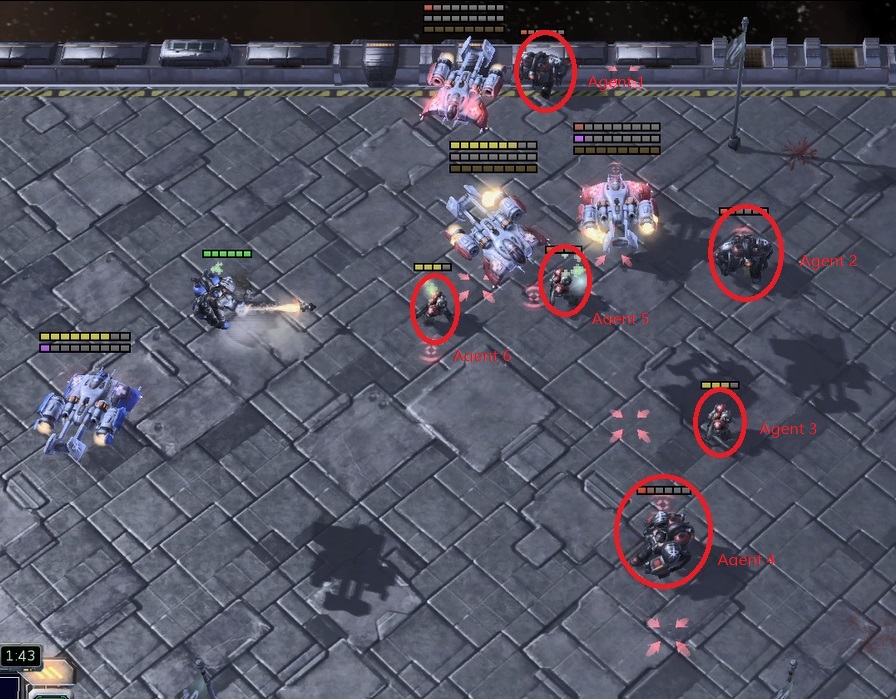}\\
            (c) & (d) 
\end{tabular}
\caption{
Visualization of the wandering issue on Defense\_3\_Subtask using a policy learned by QMIX. 
Agents exhibiting this behavior are highlighted with red circles. 
After completing their assigned atomic subtasks, these agents remain idle and wander locally instead of contributing further. 
In contrast, a successful policy should enable such agents to reallocate their efforts and assist others once their primary responsibilities are fulfilled.}
\label{fig:wandering}
\end{center}
\end{figure}

In \textbf{Defense\_3\_Subtask}, CDS maintains a constant episode length of 7 (Fig.~\ref{fig:epLength}(a2)), indicating that it fails at least one defense subtask and fails to learn a DOL policy.
LDSA achieves episode lengths exceeding 7 but shows a downward trend over time, suggesting an awareness of DOL without learning a sufficiently effective DOL policy to ensure task completion.
In contrast, other baselines exhibit steadily increasing episode lengths that eventually converge to values significantly higher than 7 (Fig.~\ref{fig:epLength}(a1,a2)), indicating stronger DOL learning capacity than CDS and LDSA.
A similar trend is observed in \textbf{Defense\_4\_Subtask} (Fig.~\ref{fig:epLength}(b1,b2)).
However, one notable distinction is that both QMIX and ROMA exhibit stable episode lengths at 7, implying that they fail to learn any effective DOL policy throughout training.
Furthermore, the converged episode lengths of most baselines decrease from Defense\_3\_Subtask to Defense\_4\_Subtask, aligning with the intentional increase in task complexity by design.
In \textbf{Mixed\_2\_Subtask}, all baselines except RODE maintain an episode length of 21 (Fig.~\ref{fig:epLength}(c1,c2)), suggesting that they fail to complete the pursuit subtask.
RODE shows a consistent increase in episode length, exceeding 60 by the end of training.
Nevertheless, its persistent test winning rate of 0 indicates failure in the defense subtask.
Upon analysis, this failure stems from the agent assigned to the pursuit subtask remaining idle and wandering after completing its objective—an issue we refer to as the \textbf{wandering issue}.
We visualize the example of the wandering issue in Fig.~\ref{fig:wandering}.
In \textbf{Mixed\_3\_Subtask}, the episode length of CDS initially increases to approximately 18 but later declines and stabilizes at 7, indicating instability and ineffectiveness in learning DOL policies.
RODE again demonstrates a steadily rising episode length, though the final convergence value is lower than in Mixed\_2\_Subtask.
ROMA exhibits stable episode lengths of 21 during early training but a sharp increase in later stages, suggesting eventual learning of a policy to complete the pursuit subtask.
However, its test winning rate remains at 0, indicating that its learned policy is insufficient for ensuring all subtasks are completed.
The episode lengths of other baselines remain stable at 21, underscoring their inability to complete the pursuit subtask (Fig.~\ref{fig:epLength}(d1,d2)).

In summary, our findings reveal that \textbf{all baselines exhibit some degree of DOL capability across all or a subset of the CTC tasks.
However, none of the baselines can learn an effective DOL policy to form successful cooperation, as evidenced by their inability to complete the tasks.}
These results underscore that CTC tasks pose significant challenges for these cooperative MARL methods, and raise concerns regarding the solvability of CTC tasks.
\textbf{Therefore, we develop a guiding solution for the CTC tasks to demonstrate solvability and facilitate further research.}

\begin{figure}
    \centering
    \includegraphics[scale=0.3]{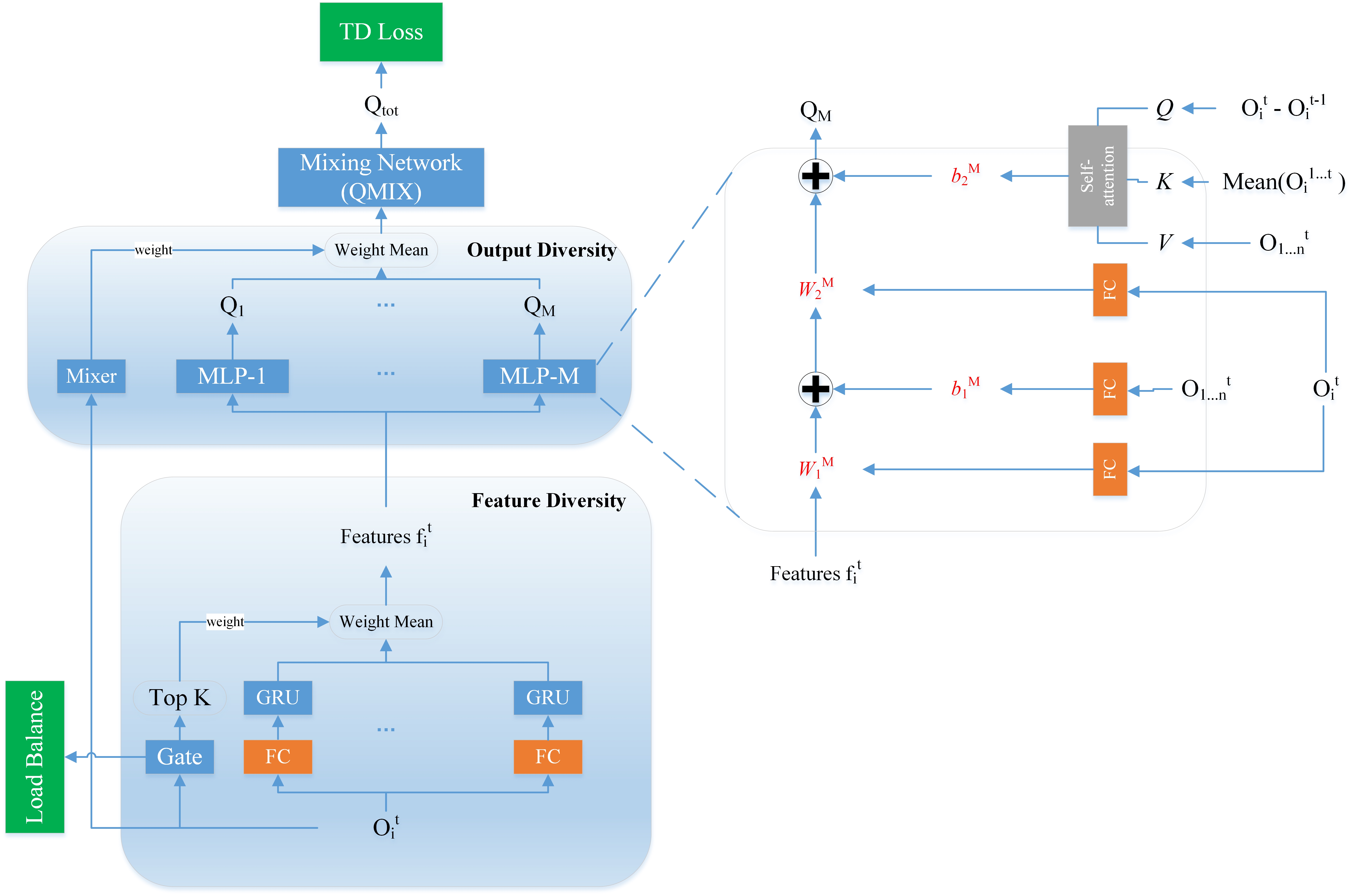}
    \caption{The policy network of extended QMIX.}
    \label{fig:eQMIX}
\end{figure}

\section{A Guiding Solution}
The guiding solution serves two primary objectives: (1) to demonstrate the solvability of the CTC tasks, and (2) to provide a guide for future research on DOL and cooperation in cooperative MARL.
Due to the high complexity of the CTC tasks, we manually design a rule-based external reward (RER) informed by domain knowledge.
To better leverage the external reward signal, we further extend the policy network of QMIX (e-QMIX), enabling improved performance and more effective learning of DOL and cooperation.

\subsection{Rule-based External Reward}
\label{rer}
\textbf{Damage Reward} Considering the high priority of the pursuit subtask, the designed reward prioritizes the completion of the pursuit subtask.
In CTC tasks, the pursuit subtask is executed through a specialized unit, Medivac, which plays a pivotal role in both the pursuit and defense subtasks.
This importance arises from two distinctive behavioral features.
\textbf{Feature 1}: When no allied units remain, the Medivac continues to move toward its target point, even when under attack.
\textbf{Feature 2}: When accompanied by allied units, the Medivac halts and heals them upon being attacked. Once all allies are defeated, it resumes its path toward the target, again ignoring incoming damage.
Feature 1 makes the Medivac a time-sensitive threat in the pursuit subtask.
Failure to intercept and eliminate it within a limited time window results in immediate task failure.
Feature 2 introduces a prioritization dynamic: targeting the Medivac early disrupts enemy coordination and increases the likelihood of successful subtask completion.
Therefore, prioritizing the defeat of the Medivac is a sound and effective policy across both pursuit and defense subtasks.
To incentivize this behavior, we design an extrinsic reward based on the reduction in the Medivac’s health over time (damage reward), encouraging agents to actively engage with and eliminate these high-priority targets.
\begin{equation}
    r^t_{\text{damage}} = \frac{1}{|\mathcal{M}|} \sum_{m \in \mathcal{M}} \frac{\text{HP}^{t-1}_m-\text{HP}^t_m}{\text{HP}^{\text{max}}_m},
\end{equation}
where $\text{HP}^{\text{max}}_m$ denotes the maximum health of Medivac unit $m$, while $\text{HP}^{t-1}_m$ and $\text{HP}^t_m$ represent its health at time steps $t-1$ and $t$, respectively.
The set $\mathcal{M}$ contains all Medivac units present in the task, and $|\mathcal{M}|$ is the number of units in $\mathcal{M}$.

\textbf{Action Reward}  A crucial aspect of cooperation in CTC tasks involves assisting teammates in completing their subtasks after one’s own has been completed.
However, we frequently observe \textbf{wandering issue}, which is a key factor limiting the performance of baselines.
To address this problem, we incorporate an auxiliary reward that encourages agents to choose actions that lead to positive outcomes, specifically offensive actions that generate environment-level rewards (e.g., attacking actions).
This incentive aims to keep agents engaged in meaningful behaviors throughout the episode, thereby promoting more effective cooperation beyond initial task assignments.
\begin{equation}
    r^t_{\text{action}} = \frac{1}{n} \sum_{i=1}^n \mathds{1}(a_i^t \in A^*), 
\end{equation}
where $n$ represents the total number of agents, and $A^*$ denotes the set of attack actions.
The indicator function $\mathds{1}(a_i^t \in A^*)$ equals to 1 if agent $i$ chooses an attack action at time $t$, and 0 otherwise.

Then, \textbf{Rule-based Extrinsic Reward (RER)} is formally defined as follows:
\begin{equation}
    r^t_{\text{RER}} = \alpha r^t_{\text{damage}} + \beta r^t_{\text{action}}
\end{equation}
The coefficients $\alpha$ and $\beta$ control the relative importance of each component in RER.
The term $r^t_{\text{action}}$ encourages agents to consistently select attack actions, thereby mitigating the \textbf{wandering issue}.
Since this encouragement should remain effective throughout training, the coefficient $\beta$ is set to a constant value.
On the other hand, $r^t_{\text{damage}}$ guides the agents to prioritize targeting the Medivac units.
It facilitates learning appropriate grouping policies and temporal prioritization.
However, overly strong emphasis on $r^t_{\text{damage}}$ can overshadow the influence of $r^t_{\text{action}}$, especially in later stages of training.
To balance this, the coefficient $\alpha$ is designed to decay exponentially over time, allowing the training to initially focus on prioritization while later reinforcing sustained cooperative behavior.
This composite reward structure promotes learning effective and cooperative policies by (1) prioritizing the timely elimination of high-impact targets such as Medivacs, and (2) reducing ineffective behaviors such as wandering.
Consequently, it fosters MARL methods to learn DOL and cooperation policies, which are essential to solving the CTC tasks.

\begin{figure}[t]
\begin{center}
\begin{tabular}{@{\extracolsep{\fill}}c@{}c@{}c@{}c@{\extracolsep{\fill}}}
            \includegraphics[scale=0.23]{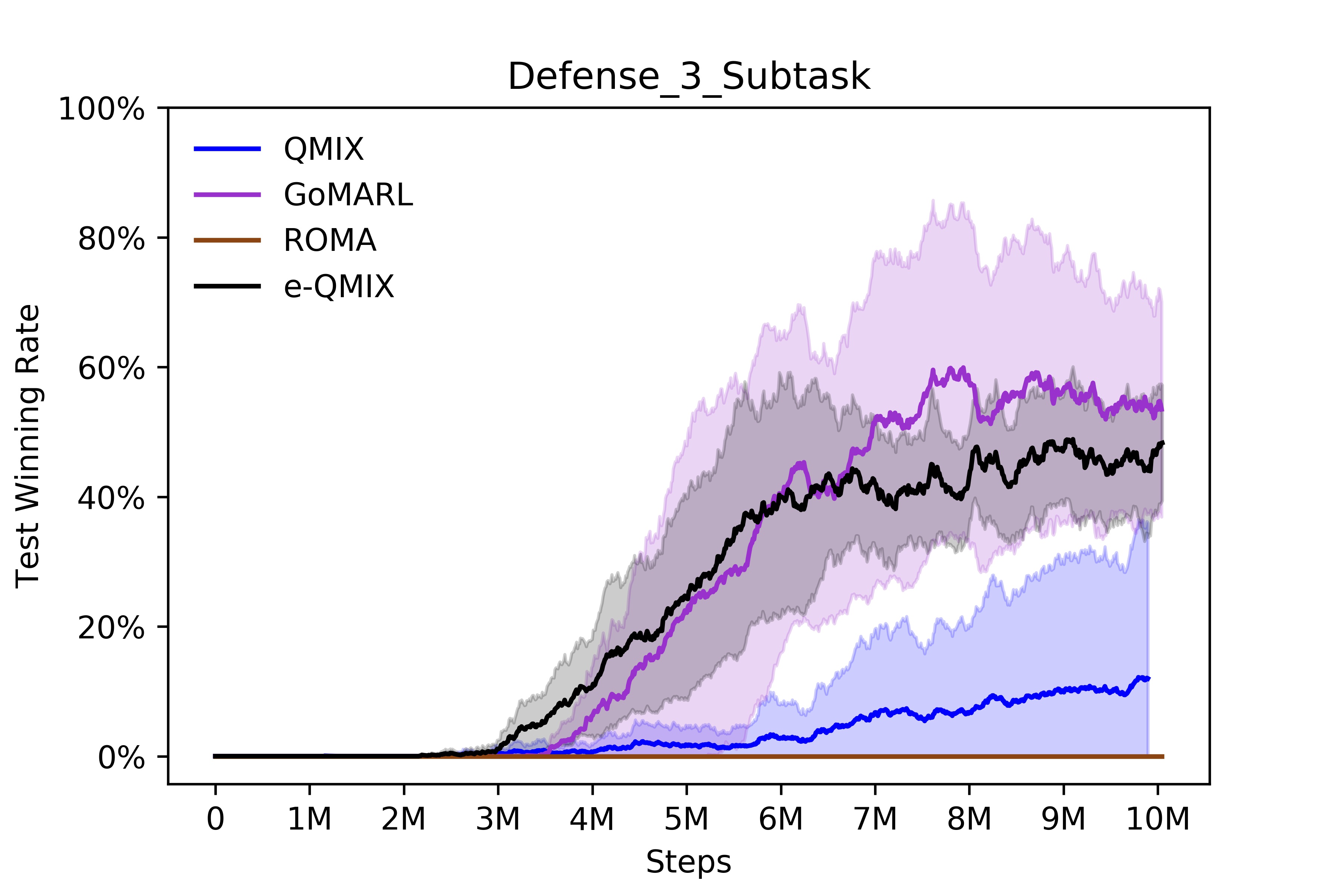} &
            \includegraphics[scale=0.23]{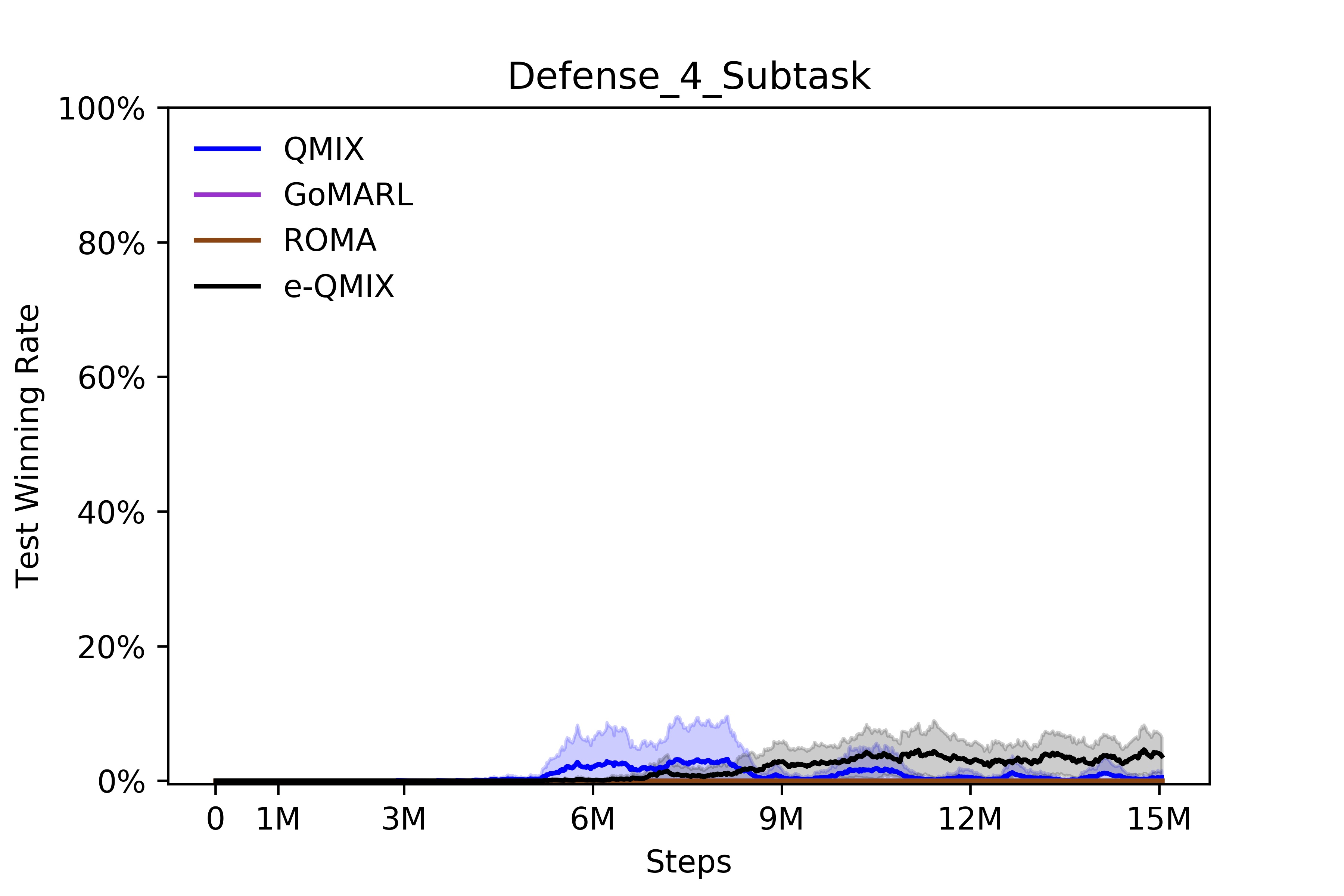} &
            \includegraphics[scale=0.23]{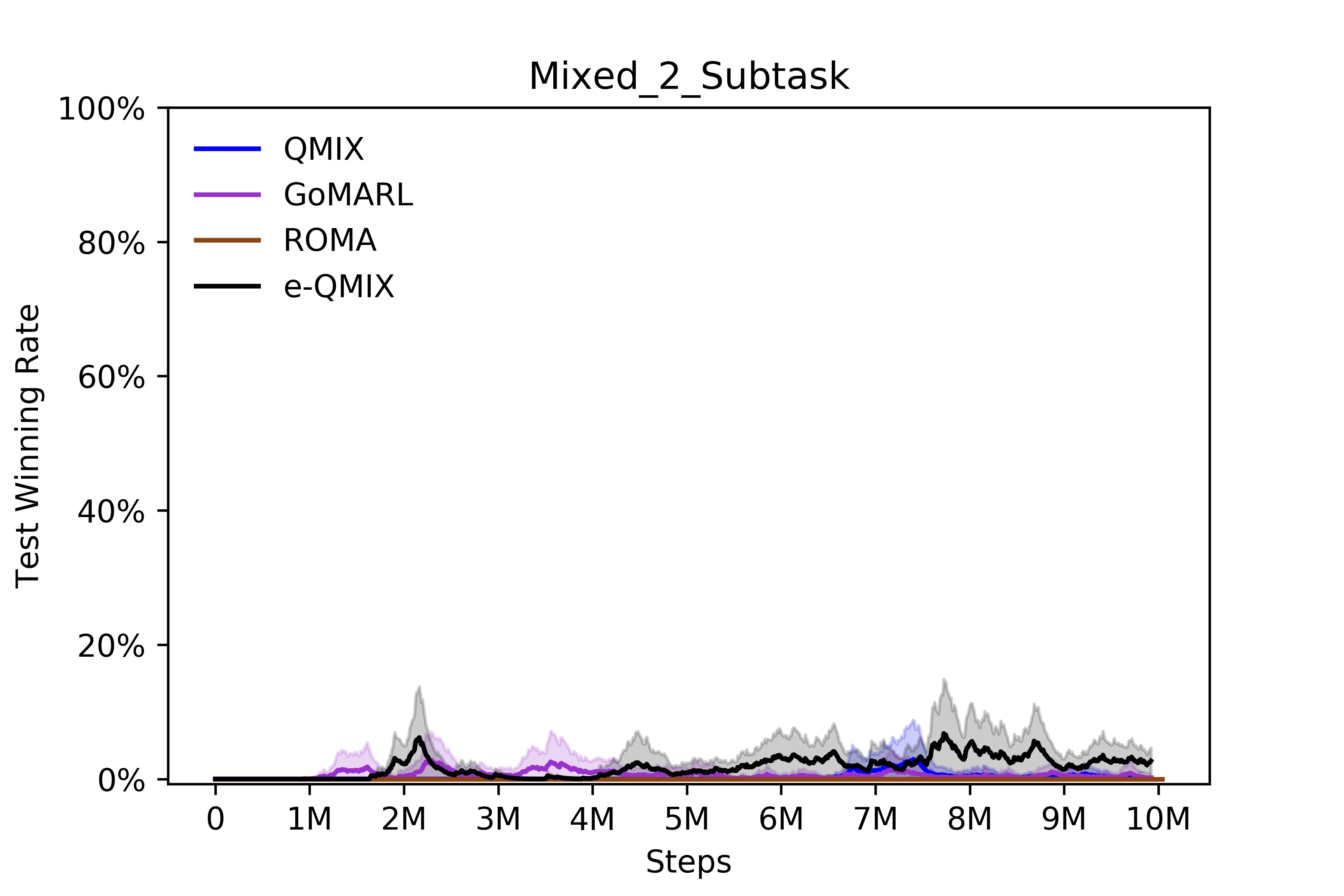} &
            \includegraphics[scale=0.23]{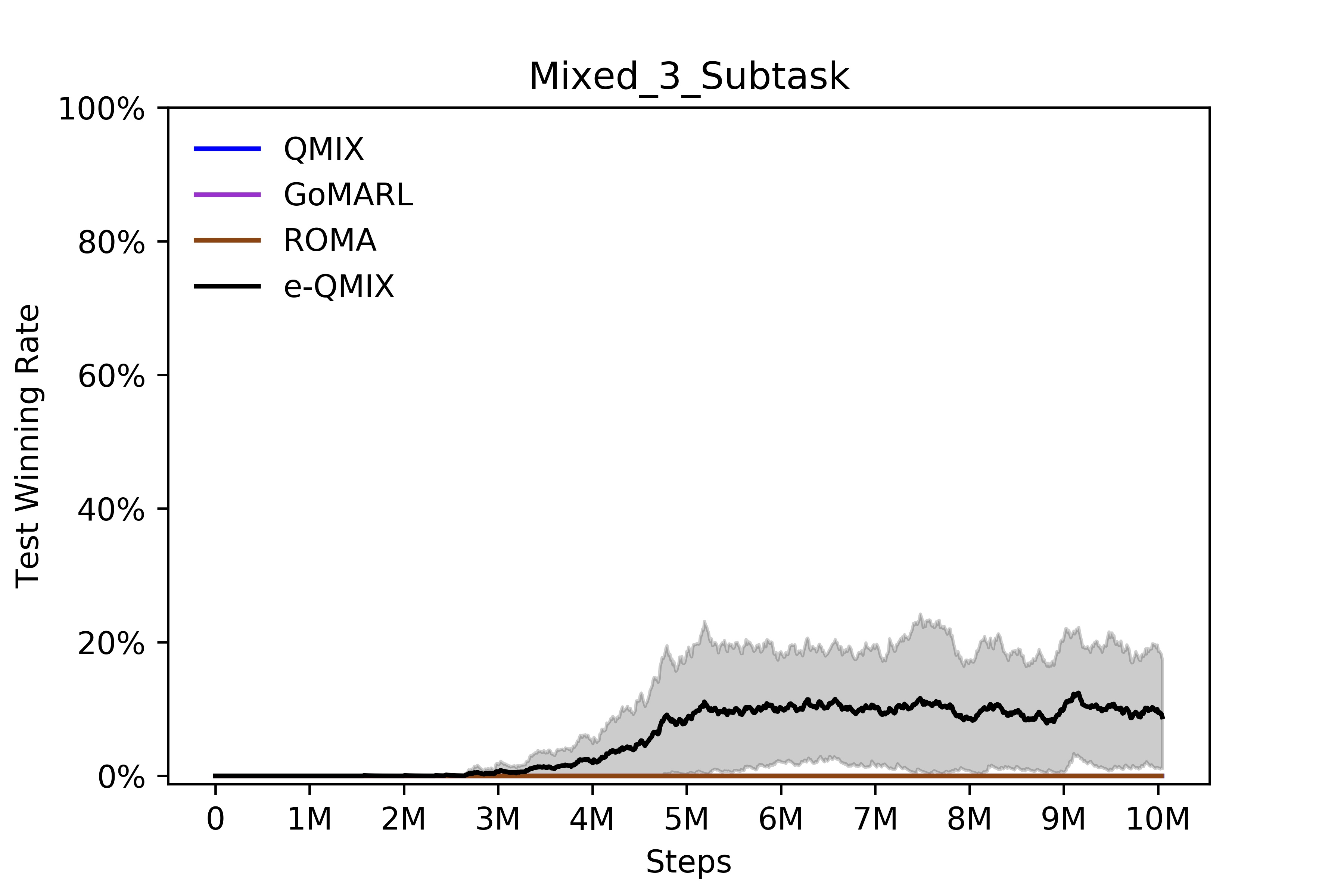}\\
            (a1) & (b1) & (c1) & (d1) \\
            \includegraphics[scale=0.23]{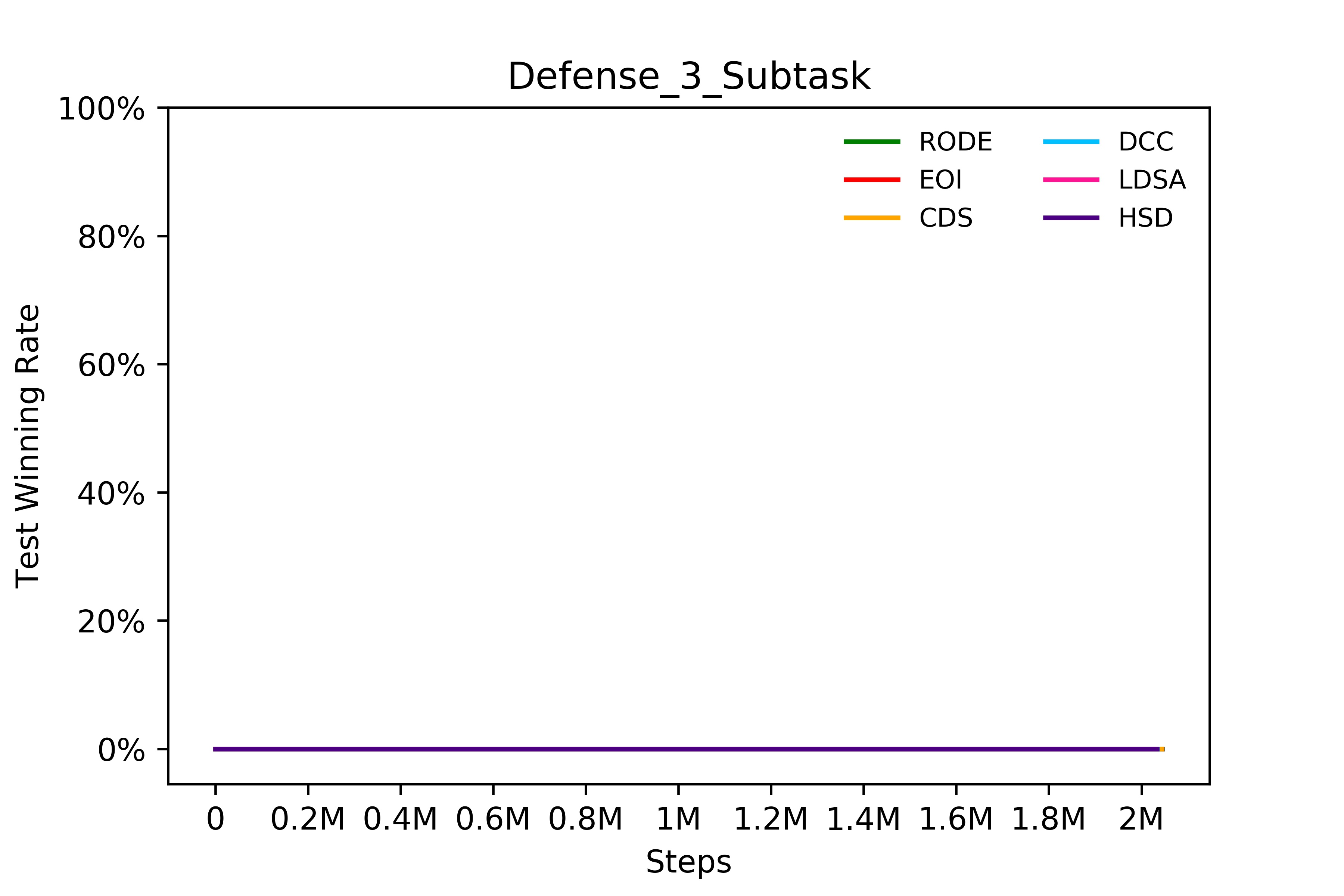} &
            \includegraphics[scale=0.23]{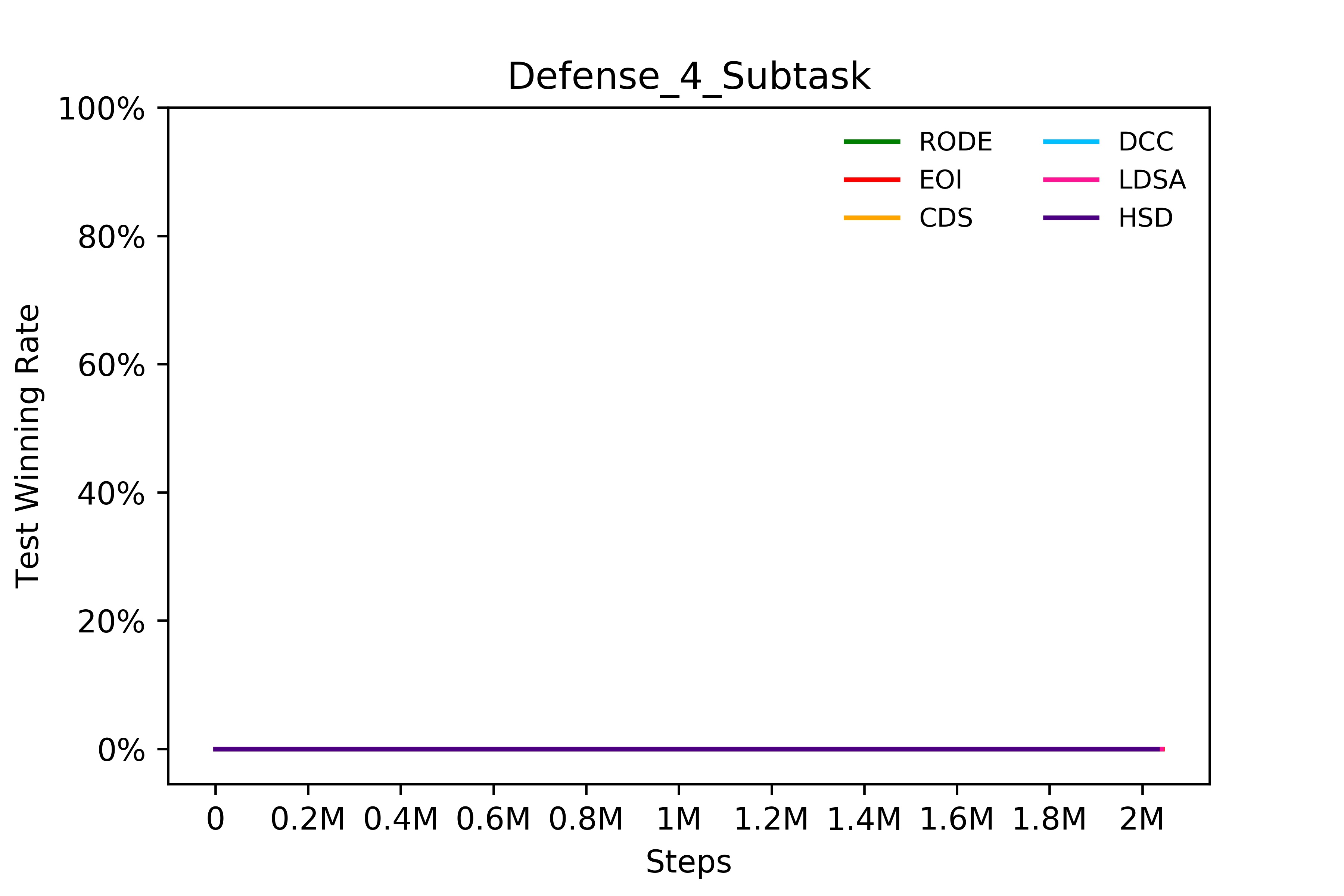} &
            \includegraphics[scale=0.23]{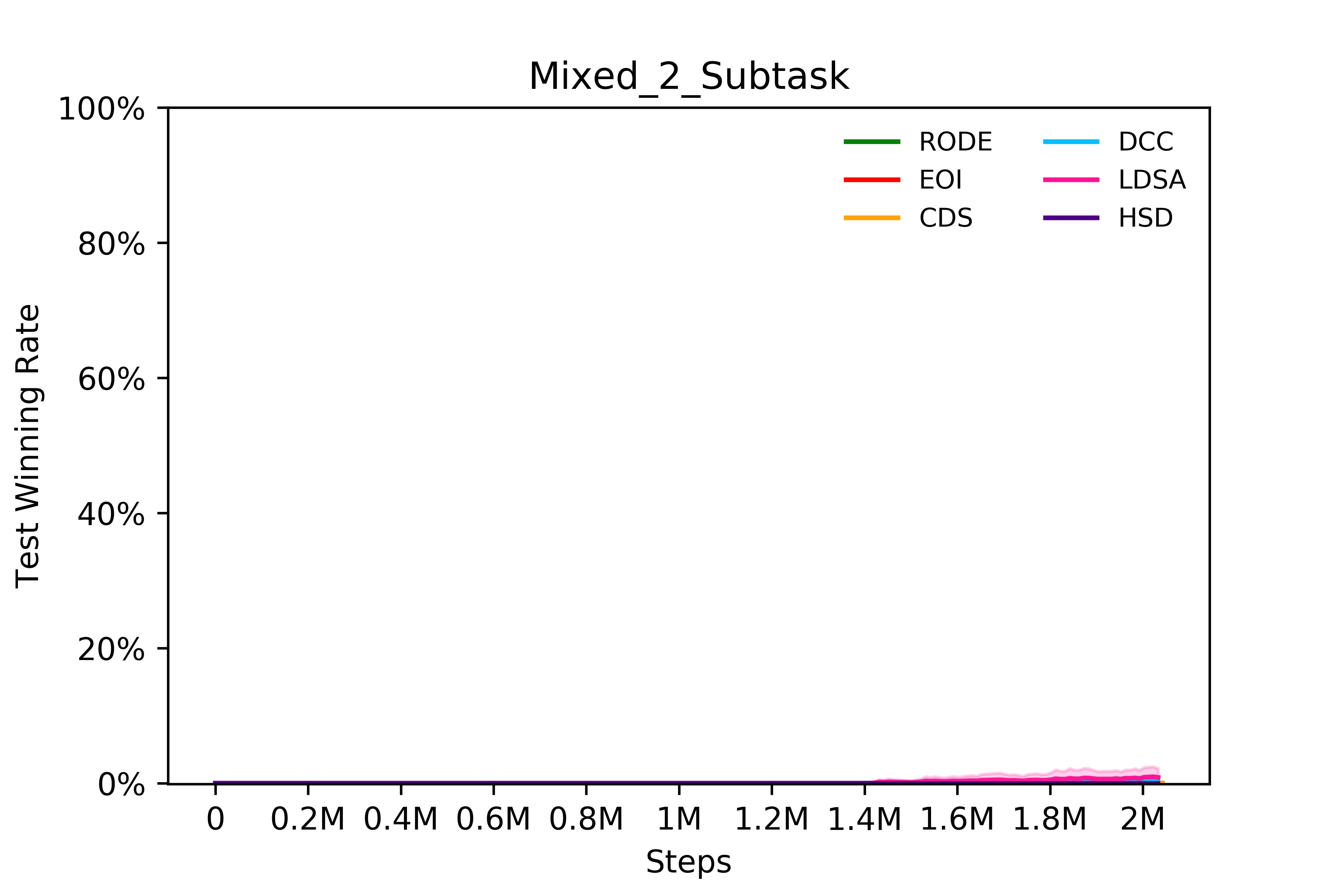} &
            \includegraphics[scale=0.23]{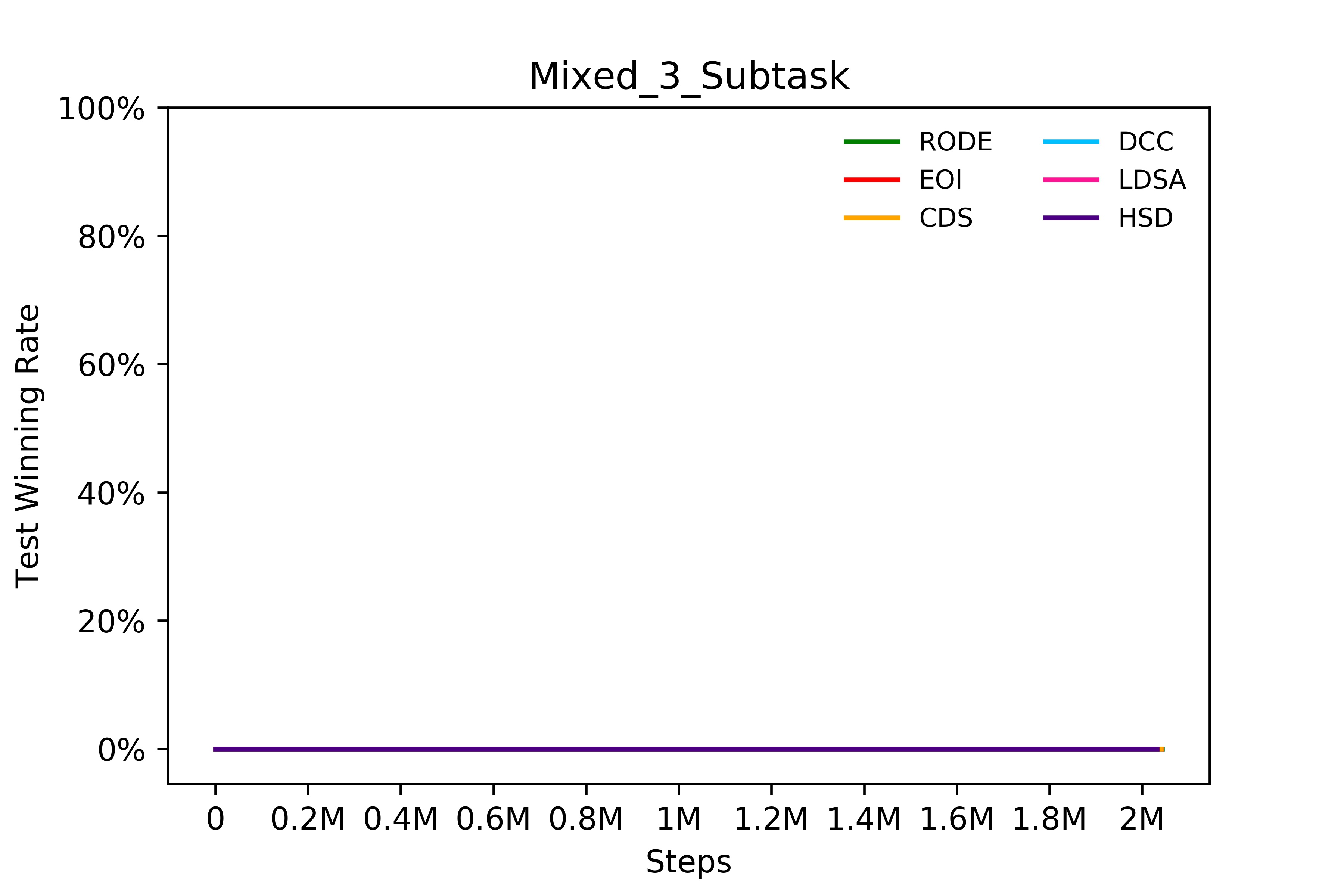}\\
            (a2) & (b2) & (c2) & (d2) \\
\end{tabular}
\caption{Performance of baselines on CTC tasks with RER.}
\label{fig:performance2}
\end{center}
\end{figure}

\subsection{Extended QMIX}
\label{sec:eqmix}
The CTC tasks demand a high degree of policy diversity among agents duo to DOL and cooperation.
Theoretically, successful completion of these tasks requires agents to form distinct groups, each assigned to specific subtask, with the additional expectation that agents provide inter-group support once their own subtasks are completed.
\textbf{To address this requirement, we extend the policy network of QMIX to enhance its capacity for learning diverse agent behaviors.}
We refer to this enhanced version as e-QMIX, and its policy network architecture is illustrated in Fig.~\ref{fig:eQMIX}.
The policy network in e-QMIX is composed of two sequential modules:
\begin{enumerate}
    \item Feature Diversity — designed to enable agents to extract differentiated features even when sharing the same network.
    \item Output Diversity — responsible for producing diverse action policies conditioned on those distinct features.
\end{enumerate}
Importantly, this extended policy network is still shared across all agents, preserving the parameter-sharing structure of QMIX while enabling richer behavior specialization necessary for solving CTC tasks.

The \textbf{feature diversity} is implemented by a sparse mixture of experts (MoE) framework~\cite{shazeer2017outrageously}.
Sparse MoE can promote diversified feature representations which is important for policy diversity.
Furthermore, sparse MoE can mitigate the convergence of agent policies that often arises from parameter sharing in multi-agent systems~\cite{li2021celebrating}, which hinders policy diversity.
\begin{align}
     f_i^t &=  \sum_{k=1}^M g_{i,k}^t E_k(o_i^t), \\
     g_{i,k}^t &= \begin{cases}
         s_{i,k}^t, \ \ \ s_{i,k}^t \in \text{TopK}(\{s_{i,j}^t|1 \le j \le M\}), \\
         0, \ \ \ \text{otherwise},
     \end{cases} \\
     \boldsymbol{s}_i^t &= \text{Gate}(o_i^t),
\end{align}
Here, $f_i^t$ denotes the feature embedding of agent $i$ at time step $t$, generated as a weighted sum of expert outputs $E_k(o_i^t)$, where $E_k$ is the $k$-th expert network and $o_i^t$ is the agent's local observation.
The gating network $\text{Gate}(o_i^t)$ produces the score vector $\boldsymbol{s}_i^t$, from which the top-$K$ scores are selected to determine the expert weights $g_{i,k}^t$.
This sparse selection ensures that each agent leverages only a subset of available experts, encouraging diverse representations.
To prevent load imbalance across experts—which can lead to under-utilization of some experts—we incorporate a balance loss~\cite{liu2024deepseek}:
\begin{align}
    \mathcal{L}_{Bal} &= \gamma \sum_{k=1}^M c_k P_k ,\\
    c_k &= \frac{1}{T} \sum_{t=1}^T \mathcal{1}(s_i^t \in \text{TopK}(\{s_j^t|1 \le j \le M\})), \\
    P_k &= \frac{1}{T} \sum_{t=1}^T s_i^t
\end{align}
where $\gamma$ is a small weighting hyperparameter, $\mathcal{1}(\cdot)$ is the indicator function, and $T$ denotes the trajectory length.
The loss encourages an even distribution of expert usage across time steps, ensuring that all experts are engaged throughout training.

The \textbf{output diversity} component is designed to generate diverse policies for agents, further enhancing behavioral differentiation.
To achieve this, we employ a mixture of experts architecture~\cite{jacobs1991adaptive}, where the final action-value estimate for each agent is computed as:
\begin{align}
    Q_i &= \sum_{k=1}^M \pi_k E'_k(f_i^t), \\
    \pi_k &= \text{Mixer}(f_i^t),
\end{align}
Here, $E'_k(f_i^t)$ denotes the output of the $k$-th expert in the output diversity module, and $\pi_k$ is the corresponding mixing coefficient obtained from a learned mixer network.
Each expert $E'_k$ consists of a two-layer fully connected (FC) network.
The weights $W_1$ and $W_2$ of the FC layers are dynamically generated by separate FC networks conditioned on the agent's local observation.
The bias terms are computed as follows: $b_1$ is generated by a FC that takes as input the concatenated observations of all agents.
$b_2$ is derived using a multi-head attention mechanism, where:
The query is the agent's observation change (i.e., temporal difference in observations); The key is the historical average of the agent's observations; The value is the concatenated observations of all agents.

After computing the individual agent Q-values $Q_i$, the subsequent procedure mirrors that of QMIX: each $Q_i$ is passed through a mixing network to form a total Q-value used for computing the TD loss, which then drives backpropagation for policy optimization.
This structure promotes both representational and decision-level diversity, essential for effective DOL and cooperation.

\subsection{Performances}
The performance of each baseline with RER is presented in Fig.~\ref{fig:performance2}.
First, e-QMIX consistently achieves non-zero test winning rates across all CTC tasks, thereby demonstrating the solvability of the CTC tasks.
Second, RER significantly enhances the performance of both GoMARL and QMIX in certain tasks.
For instance, in Defense\_3\_Subtask (Fig.~\ref{fig:performance2}(a1)), RER markedly boosts GoMARL's performance, yielding a maximum test winning rate exceeding 80\%. QMIX also benefits from RER, achieving a stable average test winning rate above 10\%.
Supported by RER, e-QMIX performs comparably to GoMARL in this task.
In Defense\_4\_Subtask (Fig.~\ref{fig:performance2}(b1)), the performance gains from RER are less pronounced.
Only QMIX shows marginal and unstable improvements, while e-QMIX continues to achieve a stable non-zero average test winning rate.
A similar trend is observed in Mixed\_2\_Subtask (Fig.~\ref{fig:performance2}(c1)), where e-QMIX maintains a stable average test winning rate above zero, while QMIX and GoMARL do not exhibit consistent improvements.
In Mixed\_3\_Subtask (Fig.~\ref{fig:performance2}(d1)), QMIX and GoMARL maintain average test winning rates near zero.
The e-QMIX, however, achieves a modest test winning rate around 10\%, albeit with substantial variance across seeds.
Interestingly, e-QMIX performs better on Defense\_3\_Subtask than on Defense\_2\_Subtask, which may appear counterintuitive given that Defense\_3\_Subtask is designed to be more difficult.
This phenomenon can be attributed to the increased number of agents in Defense\_3\_Subtask, which offers higher fault tolerance and facilitates more flexible DOL.
Nevertheless, RER does not improve ROMA’s performance in any of the four tasks.
A similar outcome is observed for the baseline using the episode runner (as shown in Fig.~\ref{fig:performance2}(a2-d2)), further underscoring the non-universality of RER, as it is not equally effective across all MARL methods.
% More experimental results are provided in Sec.~\ref{sec_ers}.

\begin{figure}
\begin{center}
\begin{tabular}{@{\extracolsep{\fill}}c@{}c@{\extracolsep{\fill}}}
            \includegraphics[scale=0.25]{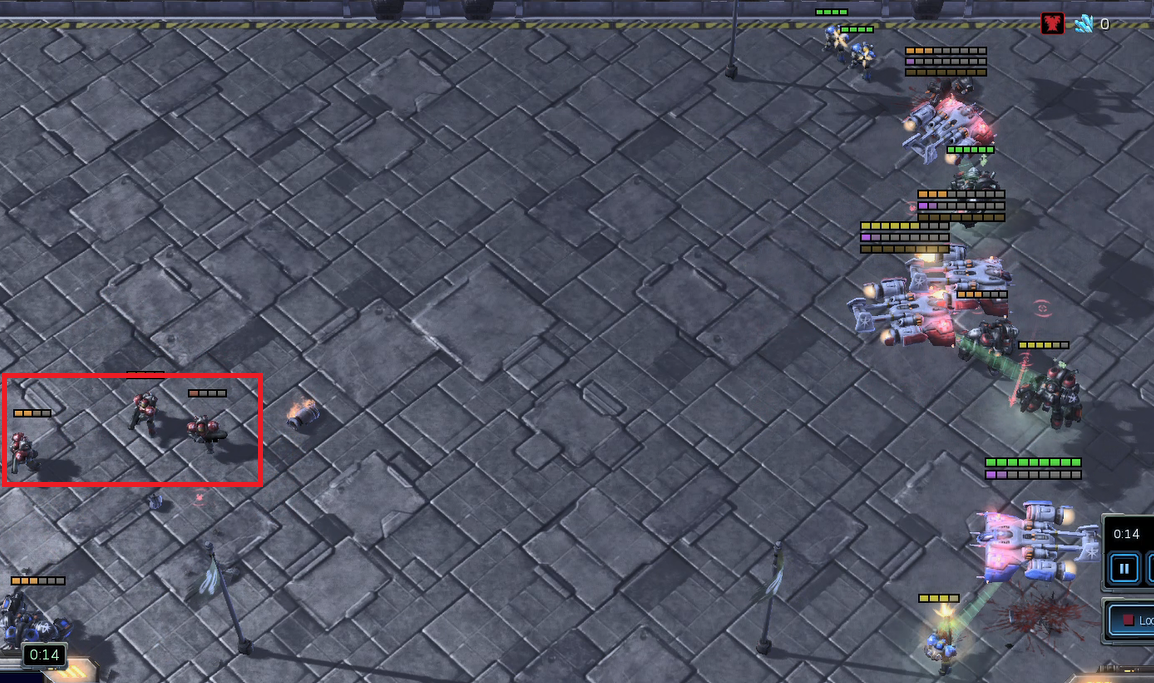} &
            \includegraphics[scale=0.25]{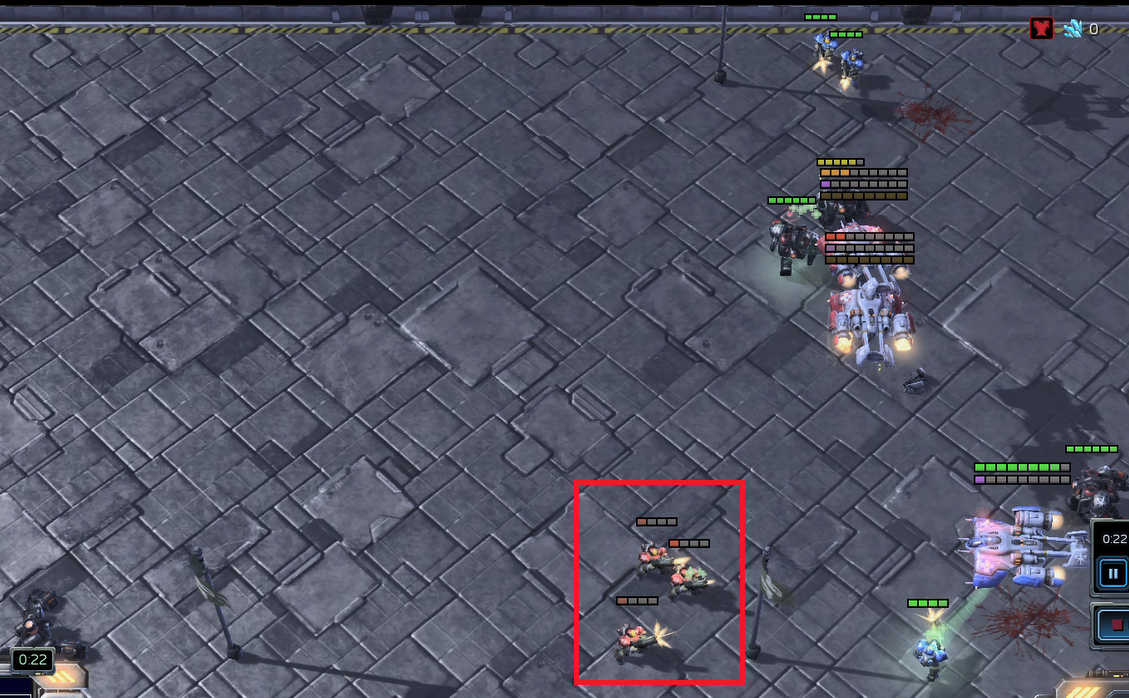} \\
            (a) & (b) \\
\end{tabular}
\caption{Visualization of Defense\_3\_Subtask. The result is obtained by the policy learned by e-QMIX. We marked the agent \textbf{without} the wandering issue with a red rectangle.}
\label{fig:solve-wandering}
\end{center}
\end{figure}

In summary, \textbf{with the support of RER, e-QMIX demonstrates the ability to learn a DOL policy and achieve successful cooperation across all CTC tasks}.
The strong performance of e-QMIX confirms the solvability of CTC tasks.
And we also show the example that e-QMIX solves the wandering issue in Fig.~\ref{fig:solve-wandering}.
However, RER fails to improve most other baselines, indicating that it is not a universal solution.
Moreover, RER is specifically tailored to CTC tasks implemented in SMAC and may not generalize to CTC tasks instantiated in other environments.
Therefore, the development of more general and widely applicable solutions for CTC tasks remains an open challenge.
Nonetheless, e-QMIX and RER provide valuable insights and lay a foundation for future research on designing effective mechanisms for DOL and cooperation in cooperative MARL.

\section{Conclusion}    
\label{Conclusion}
In this study, we propose the CTC (\textbf{C}omposite \textbf{T}asks \textbf{C}hallenge) tasks, designed to bridge the gap between cooperative MARL methods and practical applications.
CTC tasks explicitly require the MARL methods to learn a DOL policy to form successful cooperation, addressing the lack of standardized testbeds that evaluate these core capabilities in existing MARL methods.
The CTC tasks are specifically constructed to evaluate and incentivize effective DOL mechanisms, thereby unlocking the full potential of DOL in cooperative MARL.
The design of the CTC tasks ensures that DOL are necessary prerequisites for task success, and that the failure of any individual subtask results in the overall task failing.
Implemented within the SMAC testbed, the CTC tasks inherit the rationality of SMAC while minimizing the implementation overhead for researchers.
We evaluate 9 representative cooperative MARL methods on the CTC tasks.
The results reveal that all baselines get test winning rate of 0 on CTC tasks, underscoring the difficulty and discriminative power of the CTC tasks.
To demonstrate solvability and support future research, we introduce a guiding solution including RER and e-QMIX.
Experimental results show that the effectiveness of RER exists, but not universal.
CTC tasks remain highly challenging for current MARL methods, reinforcing their relevance and value as a testbed for advancing research in cooperative MARL.
In summary, the CTC tasks offer a set of solvable yet practical tasks that require DOL and cooperation, providing a valuable platform for pushing the frontiers of cooperative MARL.

\section{Funding}
% This work was supported by STI 2030 Major Projects under Grant 2021ZD0200403 and Zhejiang Provincial Natural Science Foundation of China under Grant No.LD24F030002.
This work was supported by STI 2030 Major Projects (Grant No. 2021ZD0200403) and the Zhejiang Provincial Natural Science Foundation of China (Grant No. LD24F030002).

\section{Competing Interests}
The authors have no relevant financial or non-financial interests to disclose.


\bibliography{sn-bibliography}% common bib file

\end{document}